\documentclass[journal]{ieeeconf}
\IEEEoverridecommandlockouts  
\usepackage{graphicx}
\usepackage{amsmath}
\usepackage{amssymb}
\usepackage[acronym]{glossaries}
\usepackage{tikz}
\usepackage{tikz-3dplot}
\usepackage{array}
\usepackage{multirow}
\usepackage{hyperref}
\usepackage{cleveref}
\usepackage{url}
\usepackage{xcolor}

\usetikzlibrary{shapes,arrows,positioning,calc,patterns,fit,shadows, decorations.pathmorphing, decorations.pathreplacing, 3d}
\newcommand\VRule[1][\arrayrulewidth]{\vrule width #1}

\input{symbols.sty}

\renewcommand{\equationautorefname}{\fixit{use \textbackslash eqref for equations instead of \textbackslash autoref!}}

\newcommand{\appendixref}[1]{\hyperref[#1]{Appendix~\ref*{#1}}} 

\newcommand{\cooperative}{\textit{Cooperative}}
\newcommand{\faulty}{\textit{Faulty}}
\newcommand{\naive}{\textit{Naive}}
\newcommand{\cautious}{\textit{Cautious}}
\newcommand{\omniscient}{\textit{Omniscient}}

\newcommand{\wmax}{$\mathrm{W}_\mathrm{MAX}$}
\newcommand{\wg}{$\mathrm{W}_\mathrm{G}$}
\newcommand{\wgp}{$\mathrm{W}_\mathrm{GP}$}

\newcommand{\fmax}{$F_\mathrm{max}$}
\newcommand{\cw}{confidence weighting}
\newcommand{\CW}{Confidence Weighting}

\newcommand{\mfs}{message filtering strategy}

\newacronym{gnn}{GNN}{Graph Neural Network}
\newacronym{vae}{VAE}{Variational Autoencoder}
\newacronym{gp}{GP}{Gaussian Process}
\newacronym{mlp}{MLP}{Multilayer Perceptron}
\newacronym{cnn}{CNN}{Convolutional Neural Network}
\newacronym{aevb}{AEVB}{Auto-Encoding Variational Bayes}
\newacronym{rl}{RL}{Reinforcement Learning}
\newacronym{uav}{UAV}{Unmanned Aerial Vehicle}

\newcommand{\fixit}[1]{\textcolor{red}{#1}}
\newcommand{\mynote}[1]{\textcolor{blue}{#1}}

\title{\LARGE \bf
Gaussian Process Based Message Filtering for Robust Multi-Agent Cooperation in the Presence of Adversarial Communication
}

\author{Rupert Mitchell$^{1}$, Jan Blumenkamp$^{2}$ and Amanda Prorok$^{2}$
\thanks{$^{1}$Rupert Mitchell is with Department of Engineering, University of Cambridge. (e-mail:
        {\tt\small rmjm3@cam.ac.uk}).}%
\thanks{$^{2}$Jan Blumenkamp and Amanda Prorok are with Department of Computer Science and Technology, University of Cambridge. (e-mail:
        {\tt\small \{jb2270, asp45\}@cam.ac.uk}).}%
}

\begin{document}

\maketitle
\thispagestyle{empty}
\pagestyle{empty}


\begin{abstract}
In this paper, we consider the problem of providing robustness to adversarial communication in multi-agent systems.
Specifically, we propose a solution towards robust cooperation, which enables the multi-agent system to maintain high performance in the presence of anonymous non-cooperative agents that communicate faulty, misleading or manipulative information.
In pursuit of this goal, we propose a communication architecture based on \Glspl{gnn}, which is amenable to a novel \Gls{gp}-based probabilistic model characterizing the mutual information between the simultaneous communications of different agents due to their physical proximity and relative position.
This model allows agents to locally compute approximate posterior probabilities, or \textit{confidences}, that any given one of their communication partners is being truthful.
These \textit{confidences} can be used as weights in a \emph{message filtering} scheme,
thereby suppressing the influence of suspicious communication on the receiving agent's decisions.
In order to assess the efficacy of our method, we introduce a taxonomy of non-cooperative agents, which distinguishes them by the amount of information available to them.
We demonstrate in two distinct experiments that our method performs well across this taxonomy, outperforming alternative methods.
For all but the best informed adversaries, our filtering method is able to reduce the impact that non-cooperative agents cause, reducing it to the point of negligibility, and with negligible cost to performance in the absence of adversaries.
\end{abstract}

\begin{keywords}
Robustness, Multi-Agent Systems, Adversarial Communication, Deep Learning, Gaussian Processes
\end{keywords}

\section{Introduction}
Many real-world problems require the coordination of multiple autonomous agents~\cite{hyldmar_2019,nguyen_2018}. In fully decentralized systems, agents not only need to know how to cooperate, but also, how to \textit{communicate} to most effectively coordinate their actions in pursuit of a common goal. However, control policies and communication protocols designed to optimally coordinate the behavior of multiple agents are inherently fragile with respect to unexpected behavior from any one of these agents, unless they are specifically designed to be robust. 
This fragility has been observed in the case of failures or malfunctions~\cite{bjerknes_2013, saulnier_2017}, in the case of purposeful (adversarial) attacks~\cite{gil_2017, blumenkamp_2020}, as well as in worst-case scenarios, such as byzantine collusion~\cite{dolev_1982, strobel_2018}.

While effective communication is key to successful cooperation, it is far from obvious \emph{what information is crucial to the task, and what must be shared among agents.} This question differs from problem to problem and the optimal strategy is often unknown. Hand-engineered coordination strategies often fail to deliver the desired performance, and despite ongoing progress in this domain, first-principles based solutions still tend to scale poorly with growing agent team sizes, and require substantial design effort. Recent work has shown the promise of \Glspl{gnn} to learn explicit communication strategies that enable complex multi-agent coordination~\cite{foerster_2016,li_2020,tolstaya_2020,khan_2020}. 
\Glspl{gnn} exploit the fact that inter-agent relationships can be represented as graphs, which provide a mathematical description of the network topology. 
In multi-agent systems, an agent is modeled as a node in the graph, the connectivity of agents as edges, and the internal state of an agent as a graph signal. 
The key attribute of \Glspl{gnn} is that they operate in a localized manner, whereby information is shared over a multi-hop communication network through explicit communication with nearby neighbors only, hence resulting in fully decentralizable policies.

While impressive results have been achieved, these multi-agent learning approaches \emph{assume full cooperation}, whereby all agents share the same goal of maximizing a shared global reward. 
There is a dearth of work that explores whether agents can utilize machine learning to synthesize communication policies that are robust to \textit{non-cooperative} or even \textit{adversarial} communications from their neighbors.

In this work, we consider the case of \textit{learned} communication in addition to \textit{learned} control policies.
We assume that the purpose of communication is the aggregation of observations across agents and therefore, without further loss of generality with respect to the capabilities of the overall communication architecture, we restrict messages to be learned (and likely lossy) encodings of an agent's local observations.
To the extent that messages convey information about the global world, an agent can then examine this information and potentially \textit{disregard} a message if it contains information that is incongruous with its expectations of the state of the environment it is operating in. 

While there is a rich body of work on encoding-based anomaly detection (e.g.,~\cite{richter_2017, soelch_2016, utkin_2016, ribeiro_2018, wang_2020}), the aforementioned setting distinguishes our work. Instead of detecting individual anomalous observations, we are interested detecting anomalous observations within sets of observation encodings that are spatially inter-related. Specifically, \textit{we expect the encodings communicated by spatially proximate agents to share substantial mutual information, and that the structure of this non-independence will itself be dependent on the spatial arrangement of these agents in the world they are observing.}

We therefore propose a communication model, in which each agent transmits an encoded version of its local observations generated by a neural network-based auto-encoder which learns to represent these local observations in an unsupervised manner.
Leveraging the \Gls{aevb} framework introduced in~\cite{kingma_2013}, we co-optimize this auto-encoder with a joint prior distribution over the encoded observations of arbitrary sets of spatially proximate agents in order to successfully capture the expected mutual information between encodings.
For this purpose we use a \Gls{gp} with a learnable kernel function to represent a distribution over functions from agent position to encodings, functions whose values we take to have only been indirectly observed at the finite set of points which happen to each be the physical location of an agent.

We additionally propose a method of exploiting this probabilistic model of encodings, and therefore also messages, to construct a mechanism for detecting adversarial communication.
In particular, once we express a prior belief about the range of possible messages a malfunctioning or adversarial agent might send in terms of our probabilistic model, we can compare hypotheses regarding which nearby agents $j$ of some specific agent $i$ are communicating accurate information about their local region of the world, and thereby assign posterior probabilities (subjective to agent $i$) of truthfulness to each agent $j$.
These subjective posterior probabilities, which we refer to as \textit{confidences}, can be used as weights in an attention-like modification (e.g.,~\cite{velickovic_2018}), of any \Gls{gnn} layer that uses sum aggregation.
The resulting modified layer performs identically to the unmodified version in the limit of high confidences of agent $i$ in every one of its neighbors, but completely negates the impact of the communication of an agent $j$ on the output of the layer for agent $i$ in the opposite limit of low confidence of agent $i$ in $j$.

\textbf{Related Work.} \:
There is a recent body of work addressing multi-agent robustness from the perspective of robust consensus in the presence of non-cooperative agents~\cite{saulnier_2017,guerrero_bonilla_2017, saldana_2017}.
These methods provide guarantees of robustness up to some maximum number of non-cooperative agents subject to the condition of sufficient redundant connectivity (defined by $(r,s)$-robustness metrics~\cite{usevitch_2017, saulnier_2017, guerrero_gonilla_2020}) in the agents' communication network.
The key difference between these methods and ours is that they aim to robustly reach a consensus on specific global values based on local information. We generalize this approach, as the class of robust communication problems we consider do not necessarily involve the explicit estimation of any such global values.
Outlier-robust estimation algorithms such as RANSAC~\cite{fischler_1981, antonante_2020, yang_2020} are inapplicable to our class of problems for similar reasons.

We identify two broad classes of previous work applicable to our class of problems.
Some authors focus on optimizing the collective control policy itself for robustness to communication or hardware failure~\cite{schlotfeldt_2018, zhou_2020}, while others focus on the detection of malicious communication, either by dynamic watermarking, physical fingerprinting or inconsistencies between reported location and directional variations in wireless signal strength~\cite{gil_2017, renganathan_2017, porter_2020}.
None of these methods exploit the same source of information as ours: \textit{inconsistencies between the semantic implications of received messages}, and so while they may be complementary to our approach they are fundamentally different in character.

\textbf{Contributions.} \:
Our contributions are fourfold:

\textit{(i)} We introduce a novel probabilistic model of the observations of multi-agent systems which captures the position-dependent mutual information between the observations received by spatially proximate agents.
This model incorporates a learned encoding of the observations of an individual agent, and so is inherently also a model of the mutual information between these encodings, which are suitable for use as messages with a learned communication channel such as a \Gls{gnn}.

\textit{(ii)} We introduce a taxonomy of adversarial agents in order to characterize the breadth of the adversarial communication problem. Specifically, we distinguish between adversaries with different levels of knowledge about the cooperative agents, since an optimized attack varies substantially in strength depending on what information was used in its optimization.

\textit{(iii)} We use our probabilistic model to construct a message filtering strategy that uses \emph{confidence weights} for erroneous or malicious communication in a networked multi-agent system and integrate this mechanism with a standard \Gls{gnn} communication channel via an attention-like modification.
The modified \Gls{gnn} layer is capable of mitigating or entirely negating the impact of the communication of adversaries from across our taxonomy on the behavior of the overall system.
Importantly, our \cw{} mechanism imposes very low or no performance costs in the absence of such adversarial communication.

\textit{(iv)} We demonstrate these innovations in two distinct experiments. \textit{(1)} The first experiment considers a static group classification problem chosen for its complex position-based inter-agent observation correlations. \textit{(2)} The second experiment deals with a cooperative multi-agent reinforcement learning scenario where a non-cooperative agent defecting against the group can attempt to manipulate them for its own purposes.

\section{Problem Statement}
\label{sec:problem}

We consider a set of $N$ agents in 2D planar space.
The agents can communicate with each other via a network which can be described by a dynamic graph $\mathcal{G}(\mathcal{V}, \mathcal{E}(t))$ where our set of agents $\mathcal{V}$ form its vertices (i.e. $|\mathcal{V}| = N$) and its edge set $\mathcal{E}(t) \subseteq \mathcal{V} \times \mathcal{V}$ consists of ordered pairs $(v_i, v_j)$ for every such pair of agents for which $v_i$ can communicate directly with $v_j$, and can in general can vary over time $t$.
We consider each agent to have a communication radius $r_c$ and to be able to communicate directly with any other agent within this radius, i.e. $(v_i, v_j) \in \mathcal{E}(t) \Leftrightarrow ||\mathbf{x}_{ij}(t)|| \leq r_c$
where $\mathbf{x}_{ij}(t) = \mathbf{x}_{j}(t) - \mathbf{x}_i(t)$
and $\mathbf{x}_i(t)$ is the two-dimensional position vector of agent $i$ in the world at time $t$.
An agent's neighborhood $\mathcal{N}_i(t) = \{v_j: (v_i, v_j) \in \mathcal{E}(t)\}$ is then the set of all agents within its communication radius, including itself, at a given time.
We additionally assume a agent $i$ to have access to the position relative to it $\mathbf{x}_{ij}(t)$ of any agent $j \in \mathcal{N}_i(t)$ in its neighborhood.

At time $t$ each agent $i$ receives a set of observations $\mathbf{o}_{i}(t)$ of its local region of the world in the form of a vector of real numbers.
It constructs a message $\mathbf{m}_{i}(t)$, also a real valued vector, describing these observations via some encoding function $\texttt{enc}$ of its observations.
(That is, $\mathbf{m}_{i}(t) = \texttt{enc}(\mathbf{o}_{i}(t))$.)
It communicates this message to every other agent in $\mathcal{N}_i(t)$ and receives a message $\mathbf{m}_j(t)$ for every agent $j \in \mathcal{N}_i(t)$.
It then calculates a set of features $\mathbf{f}_i(t)$ from these received messages by some aggregation function $\texttt{agg}$.
(That is, $\mathbf{f}_i(t) = \texttt{agg}(\mathbf{o}_i(t), \{\mathbf{m}_j(t): j \in \mathcal{N}_i(t)\})$.)
Finally, it uses these features to either estimate some state of the overall world, or to determine an action to take in this world as $p_i$, via a policy function $\texttt{pol}(\mathbf{f}_i(t))$.
We assume that there is only one round of messages exchanged and, in particular, that each $\mathbf{m}_i(t)$ is not re-transmitted beyond $\mathcal{N}_i(t)$.
Since we consider only the communication at a given time instant in the rest of this work, we will suppress the time dependence of variables from here, denoting e.g. $\mathbf{x}_i(t)$ as $\mathbf{x}_i$.

We consider a networked system of $N$ agents. $N-F$ of these agents are cooperative while $F$ are non-cooperative, as defined below. We shall restrict the $N$ cooperative agents to be homogeneous (i.e. to share $\texttt{enc}$, $\texttt{agg}$ and $\texttt{pol}$), since heterogeneous agent teams are not the focus of this paper.

\textit{Definition 1 (Cooperative Agent):} A cooperative agent uses the shared cooperative encoding, aggregation and policy functions $\texttt{enc}_C$, $\texttt{agg}_C$ and $\texttt{pol}_C$ respectively. These functions are optimized to achieve some shared objective, the \textit{cooperative objective}.

A non-cooperative agent is in principle any agent which does not fit all of these conditions, however in this paper all non-cooperative agents $a$ we shall consider differ on at least the encoding function $\texttt{enc}_a$, since we are interested in the ways in which a non-cooperative agent can affect the interests of cooperative agents by misleading them (intentionally or otherwise), as opposed to communicating honestly and accurately but still taking non-cooperative actions in the world according to some non-cooperative policy $\texttt{pol}_a$.

\textit{Definition 2 (Robust Message Aggregation):} A particular choice of cooperative aggregation function $\texttt{agg}_C$ is considered robust with respect to a non-cooperative agent $a$ if $a$
does not cause the cooperative agents to perform less well at their cooperative objective via its messages $\mathbf{m}_a$.

\textit{Problem 1 (Robust Multi-Agent Cooperation):} Given a networked system of $N - F$ cooperative agents and $F$ anonymous, unknown non-cooperative agents, learn a robust message aggregation mechanism that mitigates the negative impact of messages sent by the $F$ non-cooperative agents.

By the $F$ non-cooperative agents being `unknown' we mean that the behavior of the non-cooperative agents is unavailable to train against when learning the cooperative aggregation mechanism.
By them being `anonymous' we mean that they cannot be differentiated from cooperative agents by any means other than examining their communication.

\pgfdeclarelayer{bg}    
\pgfsetlayers{bg,main}  

\tikzset{
    block/.style = {draw, rectangle},
    cascaded/.style = {%
        general shadow = {%
            shadow scale = 1,
            shadow xshift = -0.5ex,
            shadow yshift = 0.5ex,
            draw=lightgray,
            fill = white,
            dashed
        },
        general shadow = {%
            shadow scale = 1,
            shadow xshift = -.25ex,
            shadow yshift = .25ex,
            draw=gray,
            fill = white
        },
        draw,
        fill = white
    },
    stackblock/.style = {
        block, cascaded,
    },
    tmp/.style  = {coordinate}, 
    concat/.style = {draw, circle, inner sep=0pt, minimum size=0.05cm},
    input/.style = {},
    output/.style = {},
    enc/.style = {trapezium, trapezium angle=-70, draw},
    inputs/.style = {draw, inner sep=1ex, rounded corners=0.1cm, dashed},
    pics/enc/.style args={#1}{code={
        \node [enc] (_e) {\texttt{enc}};
        \node [coordinate, above=0.3cm of _e] (_in) {};
        \node [coordinate, below=0.3cm of _e] (_out) {};
        \node [inner sep=0, below left=0.05cm and 0.1cm of _in] (_in_lbl) {$\mathbf{o}#1$};
        \node [inner sep=0, below left=-0.25cm and 0.1cm of _out] (_out_lbl) {$\mathbf{m}#1$};
        
        \draw[->](_in) -- (_e);
    }},
    pics/agent/.style args={#1}{code={
        \pic (_enc) {enc=#1};
        \node[circle, draw, fit=(_enc_e) (_enc_in_lbl) (_enc_out_lbl), inner sep=0pt] (#1_agent) {};
    }},
    relative to node/.style={
        shift={(#1.center)},
        x={(#1.east)},
        y={(#1.north)},
    },
    comm/.style = {decorate,decoration={snake, amplitude=0.4mm, segment length=3mm, post length=1mm}},
}

\def\centerarc[#1](#2)(#3:#4:#5)
    { \draw[#1] ($(#2)+({#5*cos(#3)},{#5*sin(#3)})$) arc (#3:#4:#5); }

\newcommand{\architecture}{
    \begin{tikzpicture}[auto, node distance=1.2cm and 0.5cm,>=latex']
        \node [block] (w) {$\texttt{c}_i(j)$};
        \node [block, left=0.5cm of w] (K) {$K$};
        \node [stackblock, cascaded, above of=K] (dx) {$\bbx_{ji}$};
        \node [stackblock, above of=w] (mj) {$\bbm_j$};
        \node [stackblock, right of=mj] (zsj) {$\bbz^{(s)}_j$};
        
        \node [block, right of=w] (gnn) {GNN};
        \node [left=0.1cm of K, inner sep=0] (agg_label) {\texttt{agg}};
        
        \node [block, below of=w] (pol) {\texttt{pol}};
        \node [left of=pol] (polc) {$p_i$};

        \node [inputs, fit={(dx) (mj) (zsj)}, label={[anchor=south east, rotate=90]north west:$j \in \mathcal{N}_i$}] (j_in_Ni) {};
        
        \node [block, inner sep=1ex,outer sep=0.4ex, fit={(agg_label) (K) (w) (gnn)}] (agg) {};
        
        \node[circle,draw, fit=(j_in_Ni) (agg),inner sep=0pt] (agent_i) {};

        \pic[above=1.7cm of agg] (enc_i) {enc={_i}};
        \draw[->](enc_i_e) -- (enc_i_out);
        
        \pic[below left=-1.2cm and 1.8cm of agent_i, rotate fit=100, transform shape] (agent_0) {agent={_a}}; 
        \draw[->, comm](agent_0_enc_e.south) -- (agent_i);

        \pic[above left=-0.2cm and 1.8cm of agent_i, rotate fit=70, transform shape] (agent_1) {agent={_b}}; 
        \draw[->, comm](agent_1_enc_e.south) -- (agent_i);

        \pic[below left=-0.2cm and 3.9cm of agent_i, rotate fit=107, transform shape] (agent_2) {agent={_c}}; 
        \draw[->](agent_2_enc_e) -- (agent_2_enc_out);
        
        \centerarc[dashed](agent_i.center)(150:210:4.85)

        \node[left=1.8cm of agent_i] (agent_i) {$r_c$};
        \node[below left=1.5cm and -3cm of agent_i] (agent_i) {$\mathcal{N}_i = \{a, b, i\}$};

        
        \draw [->] (dx) -- (K);
        \draw [->] (mj) -- (w);
        \draw [->] (mj) -- node [pos=0.4, below] {$\sim$} (zsj);
        \draw [->] (K) -- (w);
        
        \draw [->] (zsj) -- (gnn);
        \draw [->] (w) -- (gnn);
        \draw [->] (gnn) |- node [pos=0.3, left] {$\bbf_i$} (pol);
        \draw [->] (pol) -- (polc);
    \end{tikzpicture}
}
\begin{figure}[tb]
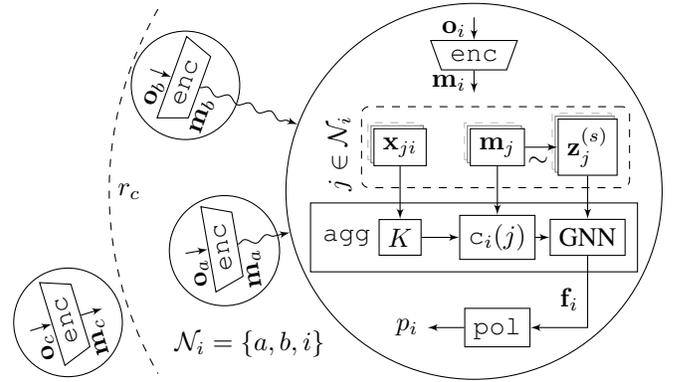

  \centering
  \architecture
  \caption{
  Schematic overview of our proposed message aggregation method from the perspective of agent $i$, represented by the enlarged circle.
  Messages $\mathbf{m}_j$ are received from every other agent (circle) inside agent $i$'s neighborhood $\mathcal{N}_i$.
  Latent vectors $\mathbf{z}_j^{(s)}$ are sampled from the posterior distributions represented by each $\mathbf{m}_j$ and combined in a \Gls{gnn} layer using attention weights calculated by a function $\texttt{c}_i(j)$ based on the full distributions and the relative positions $\mathbf{x}_{ji}$. 
  }
  \label{fig:vaegp_schema}
\end{figure}

\subsection{Assumptions}
Since there are many possible potentially harmful choices of non-cooperative behavior, goal-directed or otherwise, for any particular cooperative objective, we assume that the cooperative agents do not have any knowledge of what, if anything, the non-cooperative agent(s) are trying to achieve, let alone the particular communication strategy that the non-cooperative agents will therefore pursue. We also assume that the communication strategy of the cooperative agents is being formulated without knowing the level of knowledge that the non-cooperative agents will have about this strategy.

While the previous assumption serves to ensure generalizability of our solution, we assume additional restrictions to specify the category of solution we wish to develop.
Specifically, we focus on exploiting position-dependent correlations between agent observations to detect anomalous communication,
and we therefore place less emphasis on other potential sources of information which could be used to identify an adversarial agent.
For example, we ignore the non-communication behavior of agents,
and we only crudely check for implausibly high or low precision in communicated posteriors,
relying mostly on whether the messages are consistent with each other at their specified precision levels.
Finally, we consider only the information available in received communication from a given time instant, and do not check for inter-temporal consistency.

We assume the sole benefit of communication for the cooperative agents is aggregation of information contained in the observations $\mathbf{o}_i$ across neighborhoods.
(Note that this also constitutes substantial information about the likely actions of a neighboring cooperative agent $i$ since $\texttt{pol}_C$ is fixed and known.)
We therefore assume without further loss of generality that each agent $i$ communicates its approximate latent space posterior $\mathbf{m}_i = q(\mathbf{z} | \mathbf{o})_i$ to each member of the set of neighboring agents in its communication range $\mathcal{N}_i$.

\subsection{A Taxonomy of Non-Cooperative Agents}
\label{sec:adversary_classes}
In order to provide a more fine-grained evaluation of our method, we distinguish between four types of non-cooperative agent:
\begin{enumerate}
    \item \textit{Faulty.} The communication behavior of the non-cooperative agent is not directed towards the accomplishment of any particular goal, e.g. the agent is simply faulty.
    \item \textit{Naive.} The non-cooperative agent's behavior was optimized using knowledge of $\texttt{enc}_C$ and $\texttt{pol}_C$ as well as the effect of $\texttt{agg}_C$ on normal communication only.
    The communication behavior of the non-cooperative agent may therefore be actively manipulative, but the non-cooperative agent is unaware of, and therefore makes no attempts to evade, any manipulation detection mechanisms the cooperative agents may use.
    \item \textit{Cautious.} In addition to the information available to the \textit{Naive} agent, this agent is optimized with the knowledge that there may be some manipulation detection mechanism, but not any knowledge of its specifics. The agent therefore limits manipulative communication to lie within the range of normal communication of a cooperative agent.
    \item \textit{Omniscient.} The non-cooperative agent has perfect knowledge of the communication and detection strategies being followed by the cooperative agents and can optimize directly against them.
    The agent is exactly as manipulative as it can get away with being, given the countermeasures actually deployed in $\texttt{agg}_C$.
\end{enumerate}

Since we do not assume knowledge of the type of non-cooperative agent, we are interested in designing a method that reduces the likely impact over this broad spectrum of non-cooperative behavior on the achievement of the cooperative objective.
That is, it is desirable for the countermeasures to work as well as possible if the non-cooperative agents have low to medium knowledge of countermeasures, e.g. are \naive{} or \cautious{}, but only if this does not make the system much more vulnerable in cases of high knowledge, especially \omniscient{} agents.

\section{Preliminaries}
In this section we outline the differentiable communication channel which we build on, the \Gls{gnn}, and the inference framework we use, \Gls{aevb}, in addition to a well known instantiation of this, the \Gls{vae}.

\subsection{Graph Attention Neural Networks}
\label{sec:prel_GNN}
As described earlier in \autoref{sec:problem}, we represent the communication network of our agents as a graph $\mathcal{G}(\mathcal{V}, \mathcal{E})$, with each member of the node set $\mathcal{V}$ representing an agent. We represent the initial observations of our agents as a function whose domain is the node set $\mathcal{V}$. \Glspl{gnn} provide a means of aggregating this information across nodes while transmitting information only along graph edges and keeping all computation local to the nodes. Specifically, we consider a neural network layer of the form
\begin{equation}
    \label{eq:normal_gnn}
    \mathbf{g}_i = \sigma \left(
    \mathbf{h}_0 (\mathbf{f}_i) + 
    \sum_{j \in \mathcal{N}_i} \alpha_{ij} \: s_{ij} \: \mathbf{h}_1(\mathbf{f}_j)
    \right)
\end{equation}
where $\mathcal{N}_i$ is the neighborhood of agent $i$, $\mathbf{f}_i$ are input features and $\mathbf{g}_i$ are output features, $\mathbf{h}_0$ and $\mathbf{h}_1$ are learnable linear transforms, $\sigma$ is a pointwise nonlinearity, $\alpha_{ij}$ is an attention coefficient that indicates the importance of node $j$'s features to node $i$ \cite{velickovic_2018} and $s_{ij}$ is a normalizing constant that accounts for variability in neighborhood sizes:
\begin{equation}
    s_{ij} = \frac{1}{\sqrt{|\mathcal{N}_i| |\mathcal{N}_j|}}
    \text{.}
\end{equation}

This layer clearly does not require information from outside the communication neighborhood of agent $i$, and the computation can be done locally on agent $i$. This formulation is equivalent to a \Gls{gnn} layer~\cite{gama_2020} restricted to one communication hop between neighbors. 
The core of our work lies in the derivation of the graph self-attention coefficient $\alpha_{ij}$, which we describe in more detail in \autoref{sec:hyp_comp}. In contrast to Graph Attention Networks \cite{velickovic_2018}, we do not learn $\alpha_{ij}$, but instead calculate it with the \emph{confidence weight} function $c_i(j)$ \eqref{eq:cji} based on the full posterior latent space distributions contained in received messages $\mathbf{m}_j$ and the relative positions $\mathbf{x}_{ij}$.
These calculated attentions are shown integrated into the layer in \eqref{eq:gnn_layer} and discussed further there.

\subsection{\glsentrylong{aevb} (\glsentryshort{aevb})}
{The \Gls{aevb} framework introduces learning and inference with directed probabilistic models and, in particular, its instantiation in the form of the \Gls{vae} \cite{kingma_2013}.
Each datapoint $\mathbf{o}$ in a dataset is modeled as having been generated by a random process involving an unobserved latent variable $\mathbf{z}$.}
Specifically, $\mathbf{z}$ is first drawn from a distribution $p(\mathbf{z}; \theta)$ and then a value for $\mathbf{o}$ is drawn from $p(\mathbf{o} | \mathbf{z}; \theta)$, where the parameters $\theta$ specify a point in some space of such parameterized models.

In order to fit such a model to a dataset it is necessary to be able to estimate the marginal likelihood of the dataset according to the model specified by some point $\theta$ in parameter space.
The posterior probability $p(\mathbf{z}|\mathbf{o}; \theta)$ is unfortunately in general intractable, so they introduce a variational posterior $q(\mathbf{z} | \mathbf{o}; \phi)$ as an approximation to the true posterior, parameterized by $\phi$. This allows the marginal likelihood for a given datapoint to be stated as:
\begin{equation}
    \log p (\mathbf{o}; \theta) = D_{KL} (
    q(\mathbf{z} | \mathbf{o}; \phi) ||
    p (\mathbf{z} | \mathbf{o}; \theta)
    ) + \mathcal{L} (\mathbf{o}; \theta, \phi)
\end{equation}
where $\mathcal{L}(\mathbf{o}; \theta, \phi)$ is the evidence lower bound:
\begin{equation}
\begin{split}
    \log p (\mathbf{o}; \theta) \geq
    \mathcal{L}(\mathbf{o}; \theta, \phi) =
    & -D_{KL} (q (\mathbf{z} | \mathbf{o} ; \phi) || p(\mathbf{z}; \theta)) \\
    & + \mathbb{E}_{q (\mathbf{z} | \mathbf{o}; \phi)} \log p(\mathbf{o} | \mathbf{z} ; \theta)
\end{split}
\end{equation}
The parameters $\theta$ and $\phi$ can then be optimized to maximize the sum of $\mathcal{L}(\mathbf{o}; \theta, \phi)$ across all datapoints, as a proxy for the similar sum of $\log p (\mathbf{o}; \theta, \phi)$.

\Glspl{vae} are introduced in \cite{kingma_2013} as a simple example of this framework. They are obtained by letting $p(\mathbf{z}; \theta)$ be an isotropic Gaussian and $q(\mathbf{z} | \mathbf{o}; \phi)$ be a multivariate Gaussian of diagonal covariance structure. The distribution $p(\mathbf{o} | \mathbf{z}; \theta)$ is chosen to be either Gaussian or Bernoulli, depending on the type of data being considered.
The distribution $q(\mathbf{z} | \mathbf{o}; \phi)$ can then be seen as a probabilistic encoder, and $p(\mathbf{o} | \mathbf{z}; \theta)$ as a probabilistic decoder.

\section{Auto-Encoding Communication}
\label{sec:auto_enc_comm}
In this section we introduce our model of the communication of cooperative agents. This model allows us to express hypotheses regarding the truthfulness of a given agent in a neighborhood, so that they can be compared and, ultimately, \emph{confidence} values assigned to the agent's messages.

\subsection{Model of Cooperative Communication}

Similarly to a simple \Gls{vae}, we assume the random process generating the observations $\mathbf{o}_i$ of a particular agent $i$ to involve some hidden latent variable $\mathbf{z}_i$. We assume that each $\mathbf{o}_i$ is then drawn from a distribution $P(\mathbf{o}_i | \mathbf{z}_i; \theta)$ independently of the other latent vectors. 

We must break from the simple model of an independent \Gls{vae} for each agent because we expect that when multiple nearby agents are receiving observations of the same world, these observations will share mutual information, and therefore not be independent.
We therefore, instead of drawing each $\mathbf{z}_i$ (for each agent $i$ in the neighborhood of some arbitrary agent $j$, $\mathcal{N}_j$) independently from some position-invariant distribution $P(\mathbf{z}; \psi)$, draw all the $\mathbf{z}_i$s simultaneously from a common \Gls{gp} over position
\begin{equation}
    P(\mathbf{Z} | \mathbf{X}; \psi) = GP(\mathbf{0}, K_\psi(\mathbf{X}))
\end{equation}
where $\mathbf{Z} = [\mathbf{z}_i: i \in \mathcal{N}_j]^T$, $\mathbf{X} = [\mathbf{x}_i: i \in \mathcal{N}_j]^T$, $\mathbf{x}_i$ is the position of agent $i$, and $K_\psi$ is a kernel function generating the covariance matrices of this \Gls{gp} and is parameterized by $\psi$.
We choose a \Gls{gp} here because the starting point of a \Gls{vae} gives us latent space priors which are independently Gaussian across agents, and the most obvious refinement to introduce mutual information within a neighborhood is to let them be jointly but non-independently Gaussian across said neighborhood.
Letting the structure of this mutual information vary by spatial arrangement implies a \Gls{gp}.

In terms of the whole neighborhood $\mathcal{N}_j$ our decoder is then
\begin{equation}
    P(\mathbf{O} | \mathbf{Z}; \theta) =
    \prod_{i \in \mathcal{N}_j}
    P(\mathbf{o}_i | \mathbf{z}_i; \theta)
\end{equation}
where $\mathbf{O} = [\mathbf{o}_i: i \in \mathcal{N}_j]^T$ and each independent $P(\mathbf{o}_i | \mathbf{z}_i)$ is either Bernoulli or a diagonally covariant Gaussian depending on the observation data.
(In contrast to \cite{kingma_2013} we distinguish between the sets of parameters for $P(\mathbf{Z} | \mathbf{X})$ and $P(\mathbf{O} | \mathbf{Z})$ since we will later optimize these parameters by different methods.)
Similarly we choose the encoder (or approximate latent space posterior) to be
\begin{equation}
    Q(\mathbf{Z} | \mathbf{O}; \phi) =
    \prod_{i \in \mathcal{N}_j}
    Q(\mathbf{z}_i | \mathbf{o}_i; \phi)
\end{equation}
where each $Q(\mathbf{z}_i | \mathbf{o}_i; \phi)$, the output of a per-agent encoder, is an independent diagonally covariant Gaussian.
{A similar approach was presented in \cite{casale_2018}, which suggests a similar case of integration between \Gls{gp} and \Gls{vae}, but assumes a particularly simple form for the covariances so that the model is less expressive. The authors are interested in a substantially different application: the domain on which the functions drawn from their Gaussian Process are defined is an abstract feature space, unlike our physical space of agent positions.}

\subsection{Alternative Hypotheses for Adversarial Communication}
We shall choose $\texttt{enc}_C$ to be the probabilistic per-agent encoder of our AEVB model: $Q(\mathbf{z}_i | \mathbf{o}_i; \phi)$. Since the posterior provided by this function is a diagonally covariant Gaussian, our messages $\mathbf{m}_i$ represent it via a vector of means and a vector of standard deviations.

In order to assess the truthfulness of the messages received from neighboring agents, we compare three hypotheses regarding their origin.
While the received messages are raw posteriors over possible latent vectors $\mathbf{z}$ of unknown origin,
we assume for this analysis that they were generated by the authentic $\texttt{enc}_C$ function and express our hypotheses in terms of the origins of the observations $\mathbf{o}_i$ that could have produced them.
That is, in contrast to the likely reality, our model of general adversarial communication is therefore not a process which directly outputs arbitrary messages, but a process which generates an imaginary set of observations from some (potentially unrealistic) distribution and then encodes them faithfully.

As our null hypothesis we have:
\begin{itemize}

\item 
$H_0$ (Truth): We assume that all agents who are telling the truth have had their observations $\mathbf{o}_i$ generated by our best fit model of the world and that these observations are therefore correlated as expected based on their physical positions in the world.
\end{itemize}

Our alternative hypotheses are:
\begin{itemize}
\item
$H_1$ (Plausible Lie): This represents an agent who is telling a lie by describing a plausible scenario it could have found itself in, but which is otherwise not necessarily related to its actual surroundings.
The model is that its observation $\mathbf{o}_i$ has been generated by our best fit model of the world, but with its latent variable generated via a separate and independent draw from the \Gls{gp} over positions.
Its latent vector $\mathbf{z}_i$ is therefore uncorrelated with those of the other agents, regardless of their position or whether they are telling the truth.

\item
$H_2$ (Implausible Lie): In order to account for the fact that an agent is not restricted to telling lies that describe plausible scenarios, we need a special world model which contains those implausible scenarios which are expressible in our latent space.
In this model the random process generating the observation $\mathbf{o}_i$ is exactly the same as our best fit model, except that the latent vector $\mathbf{z}_i$ is drawn from a very high variance isotropic Gaussian instead of the best fit \Gls{gp}. (For the sake of simplicity we then take the limit of high variance and therefore obtain a uniform distribution.)
\end{itemize}

\section{Model Implementation}
In this section we describe the procedure by which our model of cooperative communication may be fitted to a dataset.
We also specify our choice of kernel function for our \Gls{gp} along with its justification.

\subsection{Model Fitting}
We have three sets of parameters: $\theta$ for our decoder, $\phi$ for our encoder and $\psi$ for our \Gls{gp}.
Analogously with \Glspl{vae}, we optimize these to maximize an approximate lower bound of the marginal likelihood of observations $\mathbf{o}_i$ observed in training conditional on agent positions $\mathbf{x}_i$ for agents $i$ in some neighborhood $\mathcal{N}_j$.
(We have many such $\mathcal{N}_j$s in our dataset and therefore maximize the sum of lower bounds on log likelihoods across the dataset.)
The following identity regarding the marginal likelihood of data can be obtained as in \cite{kingma_2013}:
\begin{equation}
\begin{split}
    \log P(\mathbf{O} | \mathbf{X} ; \theta, \psi) = 
    & D_{KL} (Q (\mathbf{Z} | \mathbf{O} ; \phi) || P(\mathbf{Z} | \mathbf{O} ; \theta, \psi)) \\
    & -D_{KL} (Q (\mathbf{Z} | \mathbf{O} ; \phi) || P(\mathbf{Z} | \mathbf{X} ; \psi)) \\
    & + \mathbb{E}_{Q (\mathbf{Z} | \mathbf{O}; \phi)} \log P(\mathbf{O} | \mathbf{Z} ; \theta)
\end{split}
\label{eq:varbayes}
\end{equation}
and since the first term is non-negative, being a KL divergence, the second and third term together represent a lower bound on the likelihood of the data.

We therefore optimize $\psi$ to minimize the following pairwise KL loss, since the kernel function it parameterizes is only guaranteed to produce valid covariances during training for pairs of agents with our implementation (see \autoref{subsec:kernel}):
\begin{equation}
    \text{KL}_\psi =
    \sum_{i \in \mathcal{N}} \sum_{j \neq i \in \mathcal{N}}
    D_{KL} (
    Q(\mathbf{Z}_{i,j} | \mathbf{O}_{i,j}; \phi) ||
    P(\mathbf{Z}_{i,j} | \mathbf{X}_{i,j}; \psi)
    )
\end{equation}
where
$\mathbf{Z}_{i,j} = [\mathbf{z}_i, \mathbf{z}_j]^T$,
and similarly for $\mathbf{O}_{i,j}$ and $\mathbf{X}_{i,j}$.
This is equivalent to treating every pair of agents in $\mathcal{N}_j$ as an independent sample from the \Gls{gp} and is therefore an approximation.

For $\theta$ and $\phi$ we define a KL loss
\begin{equation}
    \text{KL}_\phi =
    D_{KL} (
    Q(\mathbf{Z}_{\mathcal{N}_j} | \mathbf{O}_{\mathcal{N}_j}; \phi) ||
    P(\mathbf{Z}_{\mathcal{N}_j} | \mathbf{X}_{\mathcal{N}_j}; \psi)
    )
\end{equation}
and a reconstruction loss
\begin{equation}
    R = 
    -
    \mathbb{E}_{Q (\mathbf{Z}_{\mathcal{N}_j} | \mathbf{O}_{\mathcal{N}_j}; \phi)}
    \log P(\mathbf{O}_{\mathcal{N}_j} | \mathbf{Z}_{\mathcal{N}_j} ; \theta)
\end{equation}
where $\mathbf{Z}_{\mathcal{N}_j} = [\mathbf{z}_i: i \in {\mathcal{N}_j}]^T$ and similarly for $\mathbf{O}$ and $\mathbf{X}$.
We optimize $\theta$ and $\phi$ to minimize a sum (as in \cite{higgins_2016})
$\beta \text{KL}_\phi + R$
weighted by the hyperparameter $\beta$
and fall back to $\frac{\beta}{|{\mathcal{N}_j}| - 1} \text{KL}_\psi + R$ if numerical problems occur early in training due to negative determinants of the covariance matrix required to calculate
$P(\mathbf{Z}_{\mathcal{N}_j} | \mathbf{X}_{\mathcal{N}_j}; \psi)$.

\subsection{Choice of Kernel}
\label{subsec:kernel}
In order to clearly demonstrate that our method is effective without knowledge of absolute position, we implement our kernel function so that covariances are explicitly dependent only on relative position. 
With our implementation this comes at the cost of occasional inconsistent generated covariance matrices early in training. (Implementation details in \appendixref{app:kernel_implementation}.)
We also restrict the single-agent internal feature distribution to be an isotropic Gaussian, similar to a VAE.
We otherwise choose our kernel to be maximally expressive with regard to possible structures of correlation between the features of different agents.

We use a naive implementation of the calculations involving our \Gls{gp} and do not optimize its structure to maximize computational efficiency because this is unusually unimportant in our case.
This is because the main scalability challenge with \Glspl{gp} is that computation time naively scales cubically with the number of datapoints considered simultaneously, but we only ever simultaneously consider the data from within an agent's neighborhood, which is inherently limited by the communication radius $r_c$.

\section{Message Filtering for Robust Cooperation}
In this section we show how \cw{} can be introduced to the simple \Gls{gnn} layer of \autoref{sec:prel_GNN} in order to filter out inconsistent messages, and how such weights may be calculated using our model of communication from \autoref{sec:auto_enc_comm}.

\subsection{Message Filtering through \CW{}}
Each agent $i$ communicates its latent space posterior $\mathbf{m}_i = q(\mathbf{z} | \mathbf{o})_i$ to every other agent within its communication range $r_c$.
Each agent then processes this information by sampling a set of latent space features from the posteriors and using them as the input of a simple \Gls{gnn} aggregation layer, producing a local set of features $\mathbf{f}_i$:

\begin{equation}
    \mathbf{f}_i = \sigma \left(
    \mathbf{h}_0 (\mathbf{z}_i^{(s)}) + \sum_{j \in \mathcal{N}_i} \texttt{c}_{i}(j) \sqrt{\frac{1}{|\mathcal{N}_i| |\mathcal{N}_j|}} \mathbf{h}_1 \left( \mathbf{z}_j^{(s)} \right)
    \right)
    \label{eq:gnn_layer}
\end{equation}
where $\mathbf{z}_i^{(s)} \sim q(\mathbf{z} | \mathbf{o})_i$ is the sample from the posterior communicated by agent $i$, $\mathbf{h}_0$ and $\mathbf{h}_1$ are linear transformations, $\sigma$ is a pointwise nonlinearity and $\texttt{c}_{i}(j)$ is a weighting function yet to be defined (see \autoref{sec:hyp_comp}).

It can easily be seen that this formulation is equivalent to the unmodified version, \eqref{eq:normal_gnn} introduced in \autoref{sec:prel_GNN}, in the case where $\texttt{c}_{j}(i) = 1$.
We set the confidence weight $\texttt{c}_{i}(j)$ to be an approximate probability that agent $j$ is being truthful, based on all other communication received by agent $i$.
This approximate probability is calculated by comparing different hypotheses corresponding to truth or different varieties of lie as introduced in \autoref{sec:adversary_classes}, via a procedure described below in \autoref{sec:hyp_comp}.
For sufficiently high subjective confidence of agent $i$ in its neighbors $j \in \mathcal{N}_i$, all of the $\texttt{c}_{i}(j)$ then approach $1$ and we recover the initial unmodified \Gls{gnn} layer behavior.

\subsection{Computation of Confidence Weights for Messages}
\label{sec:hyp_comp}
A neighborhood hypothesis $h$ is an element of the set $\{H_0, H_1, H_2\}^{\mathcal{N}_j}$, that is, some function assigning one of the single-agent hypotheses $H_0$, $H_1$ or $H_2$ to every agent in some neighborhood ${\mathcal{N}_j}$.
The agents in ${\mathcal{N}_j}$ sharing a particular assignment according to some specific $h$ together form a subset of ${\mathcal{N}_j}$.
Three such subsets $\mathcal{G}_h$, $\mathcal{B}_h$ and $\mathcal{O}_h$ may thus be formed for $H_0$, $H_1$ and $H_2$ respectively.
The latent space prior corresponding to a hypothesis $h$ is then
\begin{multline}
    P(\mathbf{Z} | \mathbf{X}; \psi, h) = 
    P(\mathbf{Z}_{\mathcal{G}_h} | \mathbf{X}_{\mathcal{G}_h}; \psi) \times \\
    \prod_{i \in \mathcal{B}_h} P(\mathbf{z}_i; \psi, H_1) \times
    \prod_{i \in \mathcal{O}_h} P(\mathbf{z}_i; \psi, H_2)
\end{multline}
where $\mathbf{Z}_{\mathcal{G}_h} = [\mathbf{z}_i: i \in \mathcal{G}_h]$ and $\mathbf{X}_{\mathcal{G}_h} = [\mathbf{x}_i: i \in \mathcal{G}_h]$,
and we have exploited the fact that $H_1$ and $H_2$ imply observation independence from other agents.

We can then compare different hypotheses $h$
via the approximate log likelihood:
\begin{multline}
    \log P(\mathbf{O} | \mathbf{X}; \psi, \theta, h) \approx \\
    -D_{KL} (
    Q(\mathbf{Z} | \mathbf{O}; \phi) ||
    P(\mathbf{Z} | \mathbf{X}; \psi, h)
    ) + C
\end{multline}
where $C$ is constant across hypotheses (derivation in \appendixref{app:weight_derivation}). We assign priors to each neighborhood hypothesis based on priors for each single-agent hypothesis:
\begin{multline}
    \label{eq:h_log_likelihood}
    \log P(h) = |\mathcal{G}_h| \log P(H_0) + \\
    |\mathcal{B}_h| \log P(H_1) + |\mathcal{O}_h| \log P(H_2)
\end{multline}

We compare some set $\mathcal{H} \subseteq \{H_0, H_1, H_2\}^{\mathcal{N}_j}$ of such hypotheses, for example $\{ h \in \{H_0, H_1, H_2\}^{\mathcal{N}_j}: |\mathcal{G}_h| \geq |\mathcal{N}_j| - 1 \}$ if it is known that the total number of adversaries $F$ is at most one.
In combination with the log likelihood of \eqref{eq:h_log_likelihood}, we obtain a posterior probability for each hypothesis $P(h|\mathbf{O}, \mathbf{X}; \psi, \theta)$
via Bayes' theorem.
\begin{multline}
    \label{eq:h_bayes}
    P(h | \mathbf{O}, \mathbf{X}; \psi, \theta) = \\ 
    \frac{\exp \left(
    \log P(\mathbf{O} | \mathbf{X}; h, \psi, \theta) + \log P(h)
    \right)}{\exp \left(
    \sum_{\rho \in \mathcal{H}}
    \log P(\mathbf{O} | \mathbf{X}; \rho, \psi, \theta) + \log P(\rho)
    \right)}
\end{multline}
Marginalizing over $\mathcal{H}$ we then obtain the subjective probability assigned by agent $j$ to agent $i$ being truthful, that is, the probability that $i \in \mathcal{G}_h$ for whichever $h$ reflects reality.
We use this to define the confidence weight $\texttt{c}_{j}(i) = P_j(i \text{ is~truthful} | \mathbf{O}, \mathbf{X}; \psi, \theta)$ assigned by agent $j$ to $i$:
\begin{multline}
    \label{eq:cji}
    \texttt{c}_{j}(i) =
    P_j(i \text{ is truthful} | \mathbf{O}_j, \mathbf{X}_j; \psi, \theta) = \\
    \sum_{h \in \mathcal{H}_j \text{ s.t. } i \in \mathcal{G}_h}
    P(h | \mathbf{O}_j, \mathbf{X}_j; \psi, \theta)
\end{multline}
where the newly introduced subscripts $j$ re-emphasize dependence of some parameters on our initial choice of neighborhood $\mathcal{N}_j$.

Since for any single agent $P(H_0)$, $P(H_1)$ and $P(H_2)$ are assumed exhaustive and must therefore sum to 1,
we retain only $s_1 = \log P(H_0) - \log P(H_1)$ and $s_2 = \log P(H_0) - \log P(H_2)$ as tunable sensitivity parameters.

\section{Experiments}
In this section, we present two experiments demonstrating our \cw{} method.
The first experiment focuses on a cooperative perception task. Its purpose is to examine and compare the performance of our weighting method with alternatives.
The second experiment introduces the time dimension, allowing agents to take sequential decisions based on the received messages. We assess the impact of our robust message aggregation scheme on performance, when compared with non-robust baselines, as well as an alternative robust benchmark method.

\subsection{Cooperative Image Classification}
\label{sec:exp_cifar}

\tdplotsetmaincoords{70}{20}
\newcommand{\cs}{*0.2}
\def\centerarc[#1](#2)(#3:#4:#5)
    { \draw[#1] ($(#2)+({#5*cos(#3)},{#5*sin(#3)})$) arc (#3:#4:#5); }

\tikzset{
  pics/drone/.style args={#1}{code={
    \draw (0, 0, 0) coordinate (o)
        -- ++(-#1,-#1,0) coordinate (a) -- cycle (o)
        -- ++(#1,-#1,0) coordinate (b) -- cycle (o)
        -- ++(-#1,#1,0) coordinate (c) -- cycle (o)
        -- ++(#1,#1,0) coordinate (d);
    \foreach \p in {(a), (b), (c), (d)}:
        \draw \p circle[radius=0.9*#1];
  }},
  pics/cross/.style args={#1}{code={
    \draw (0, 0, 0) coordinate (o)
        -- ++(-#1,0,0) -- cycle (o)
        -- ++(#1,0,0) -- cycle (o)
        -- ++(0,-#1,0) -- cycle (o)
        -- ++(0,#1,0);
  }},
  pics/agent/.style args={#1}{code={
      \draw (0,0,0) coordinate (a) -- ++(0,\size,0) coordinate (b) -- ++(\size,0,0) coordinate (c) -- ++(0,-\size, 0) coordinate (d) -- (a);
      \coordinate (o) at (0.5*\size, 0.5*\size, 2); 
      \coordinate (g) at (0.5*\size, 0.5*\size, 0); 
      \foreach \p in {(a), (b), (c), (d)}:
          \draw[dashed] \p -- (o);
      \node[above=0.1cm of o] (agent_i) {$#1$};
      \pic[gray] (drone) at (0.5*\size, 0.5*\size, 2) {drone={0.2}};
  }},
  size/.store in=\size,
  size=9\cs,
  comm/.style = {decorate,decoration={snake, amplitude=0.4mm, segment length=3mm}},
}

\newcommand{\cifar}{
    \begin{tikzpicture}[tdplot_main_coords]
        \node[canvas is xy plane at z=0] (temp) at (0,0,0) {\includegraphics[width=6.4cm]{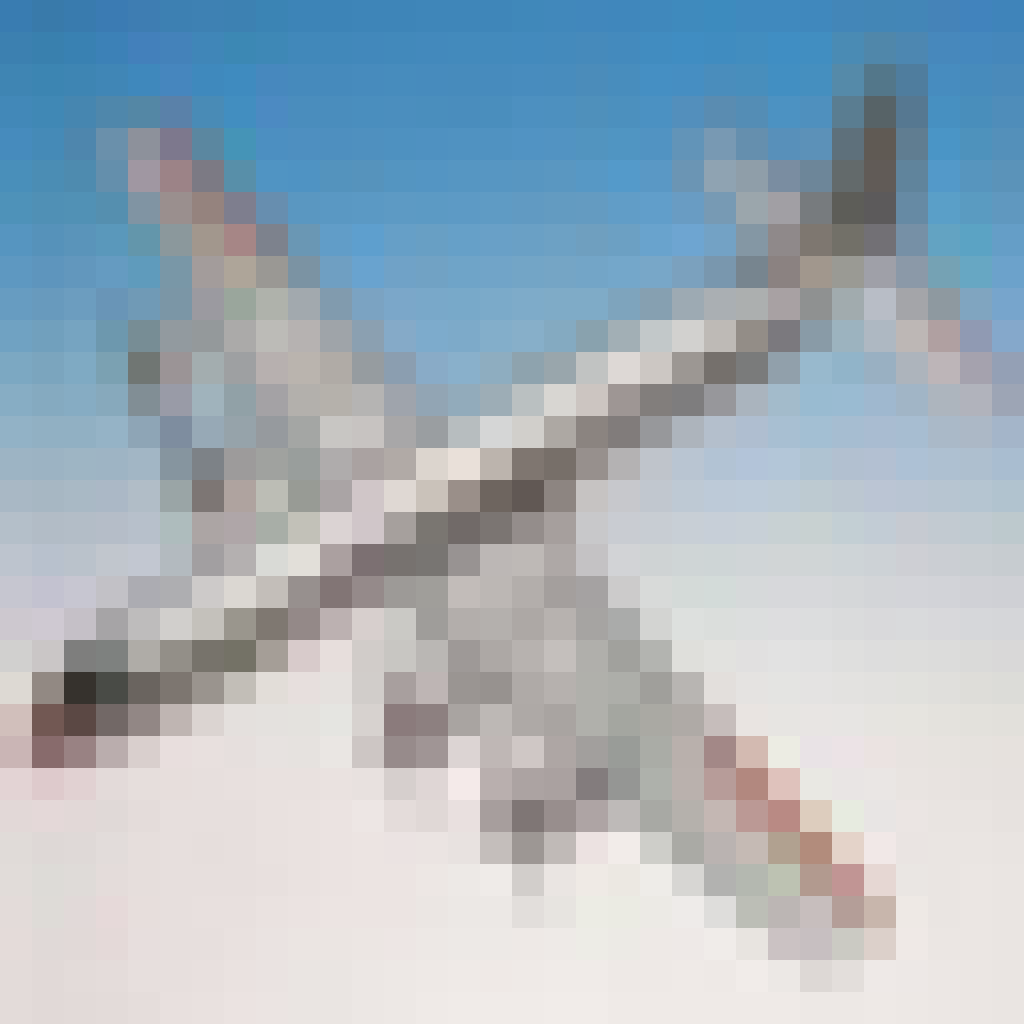}};
        \pic (aa) at (-14\cs, -13\cs, 0) {agent={a}};
        \pic (ai) at (-8\cs, -5\cs, 0) {agent={i}};
        \pic (ab) at (1\cs, -11\cs, 0) {agent={b}};
        \pic (ac) at (7\cs, 0\cs, 0) {agent={c}};
        \draw[comm] (aao) -- node [pos=0.5, above] {$\bbm_a$} (aio);
        \draw[comm] (abo) -- node [pos=0.45, above] {$\bbm_b$} (aio);

        \centerarc[dotted](aio)(160:400:13\cs)
        
        \draw[decorate,decoration={brace,mirror,raise=1pt}] (aba) -- node[left, yshift=-0.2cm]{$9$} (abd);
        \draw[decorate,decoration={brace,mirror,raise=1pt}] (abd) -- node[left, xshift=0.5cm, yshift=-0.1cm]{$9$} (abc);
        
        \node[above right=0.2cm and 0.3cm of aio] {$\ccalN_i = \{a, b, i\}$};
        \node[above left=-0.2cm and 2cm of aio] {$r_c$};
        \node at (abg) {$\bbo_b$};
        
        \draw[dotted] (aio) -- node [pos=1, right] {$\bbx_i$} (aig);
        \draw[dashed] (aig) -- node [pos=0.4, below] {$\bbx_{ai}$} (aag);
        \draw[dotted] (aag) -- node [pos=0, left] {$\bbx_a$} (aao);
        \pic (pos_ai) at (aig) {cross={0.5\cs}};
        \pic (pos_aa) at (aag) {cross={0.5\cs}};
    \end{tikzpicture}
}
\begin{figure}[tb]
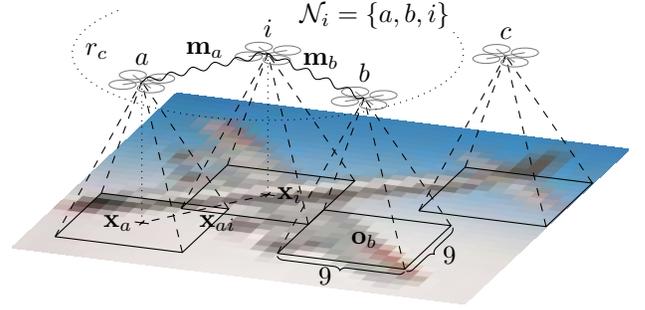

  \centering
  \cifar
  \caption{\mynote{\Gls{uav}} swarm flying over image with limited individual fields of view, resulting in an observation $\bbo_j$ for each agent $j$. The positions $\bbx_j$ can be expressed as relative positions, here visualized for agent $i$ and $a$ as $\bbx_{ai}$. Nearby \Glspl{uav} can communicate, as indicated by the waved lines.}
  \label{fig:cifar_schema}
\end{figure}

In our first case study we consider a basic multi-agent communication scenario.
The cooperative agents share their local observations with each other in order to each attempt to estimate a global categorical state.
In practice this could be a swarm of \Glspl{uav} observing the terrain they fly over with cameras and sharing their observations with each other, as illustrated in \autoref{fig:cifar_schema}.

\subsubsection{Setup}
To emulate this setup, we choose the CIFAR-10 dataset of $32\times32$ pixel images to provide our global world, with the image category being the global state to be estimated. We restrict the dataset to only the first two classes (airplanes and automobiles) for simplicity, giving us a training dataset of 10,000 images and a test dataset of 2000 images. We allow our agents to observe the local region of the image near them in a $9\times9$ grid, and distribute the $N=6$ agents uniformly across the image, subject to the condition that their vision field not extend past the image's boundary. We allow agent position to be continuous by linearly interpolating between pixel values. We also simplify the communication topology by letting the maximum communication range be infinite (leading to a fully connected communication graph).

\subsubsection{Training}
Training of the cooperative agents consists of two stages.
First we train the encoder ($\texttt{enc}_C$), decoder and kernel function to represent our dataset, using \Glspl{cnn} for the encoder and decoder.
We then train a \Gls{gnn} layer without any \cw{} and a policy layer consisting of an \Gls{mlp} using a cross-entropy classification loss.
We choose the sensitivity parameters of the \cw{} scheme $s_1$ and $s_2$ so that the mean confidence weight assigned to a cooperative agent by the weighting scheme described in \autoref{sec:hyp_comp} is $0.9$.

Our adversarial agents transform the authentic message $\mathbf{m}_C$ they would communicate if they were cooperative to the message they actually transmit $\mathbf{m}_a$ via an \Gls{mlp}\footnote{This architecture choice means that the adversary need only learn the identity function in order to begin bypassing any filtering system, since it is then indistinguishable from a cooperative agent, though harmless.}.
We also train them in a two stage process.
In the first stage we include the mean squared error of the message $\mathbf{m}_a$ with respect to $\mathbf{m}_C$ as an extra loss term in order to prevent the \Gls{mlp} from failing to find a region of the parameter space in which its messages are assigned non-negligible confidence weights.
We remove this extra term in the second phase of training.

\subsubsection{Alternative \CW{} Schemes}
We introduce two, simpler, weighting schemes for the purpose of comparison with the \cw{} scheme described in section \autoref{sec:hyp_comp}, referred  to in the rest of this section as \wgp{}.
The alternative scheme \wg{} is similar to \wgp{}, but it asks whether each message received is individually plausible while ignoring inter-agent correlations.
Since the prior distribution for a message considered independently is an zero-mean isotropic Gaussian in our model, we also introduce an even simpler scheme, \wmax{}: assign $1$ to any message with squared magnitude $< \gamma$ and $0$ otherwise, for some threshold $\gamma$.
We tune both of these methods to assign an average weight to authentic cooperative agents similar to our \wgp{} method.

\subsubsection{Results}
\begin{figure}[tb]
  \centering
  \includegraphics{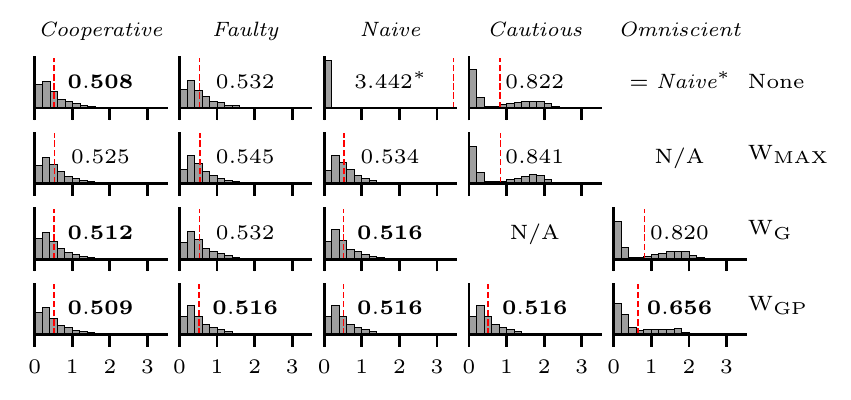}
  \caption{Cooperative loss distributions for combinations of weighting schemes and adversary knowledge classes, the mean of each loss distribution marked with a vertical red line.
  Where the definition of a knowledge class is relative to a particular weighting scheme (\cautious{}, \omniscient{}), the adversaries considered in each row are relative to the weighting scheme of that row.
  (*) Note that the distribution for the \naive{} adversary with no weighting is cropped so that only its mean is visible on the x axis, since it is substantially different in mean to the other distributions.
  Note also that the \omniscient{} adversary in the absence of any weighting is exactly the \naive{} adversary.}
  \label{fig:cifar_losses}
\end{figure}

\begin{figure}[tb]
  \centering
  \includegraphics{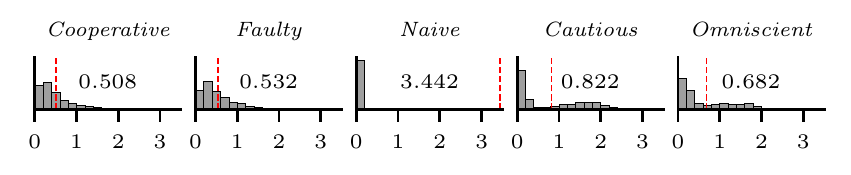}
  \caption{Loss distributions of cooperative agents in the presence of the adversaries compared against \wgp{} in the bottom row of \autoref{fig:cifar_losses}, but in the absence of a weighting scheme. The column names are relative to \wgp{}.}
  \label{fig:cifar_wnone_losses}
\end{figure}

\begin{figure}[tb]
  \centering
  \includegraphics{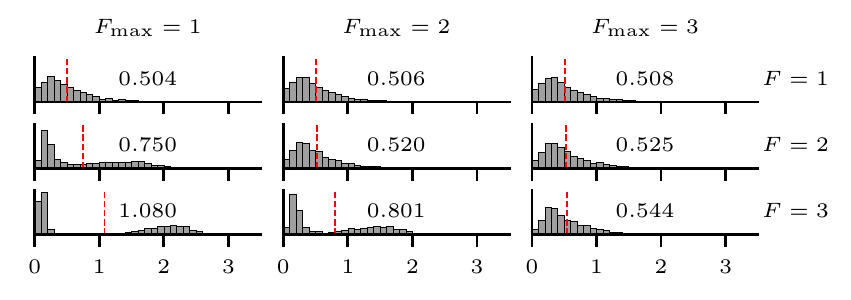}
  \caption{Cooperative loss distributions for \wgp{} in the presence of the \cautious{} adversary.
  In contrast to \autoref{fig:cifar_losses} we consider $N=8$ for this comparison.
  Each row corresponds to a choice of the actual number $F$ of adversaries present, while each column corresponds to the maximum number of adversaries \fmax{} considered possible by the weighting scheme.}
  \label{fig:cifar_losses_n8}
\end{figure}

We evaluate the performance of the \wgp{} scheme in the presence of four examples of non-cooperative agents from the knowledge classes introduced in \autoref{sec:adversary_classes}, \faulty{}, \naive{}, \cautious{} and \omniscient{} with respect to the \wgp{} scheme.
We consider a test scenario in which $F$, the number of non-cooperative agents, is either $0$ or $1$.
\autoref{fig:cifar_losses} shows the distribution and mean of the loss function of the cooperative agents, with each column corresponding to a different choice of non-cooperative agents knowledge class relative to the scheme in question, the top row corresponding to no weighting scheme, the middle rows corresponding to the alternative \wmax{} and \wg{} schemes, and the bottom row showing the \wgp{} scheme.
\autoref{tab:cifar_accuracy} shows the corresponding classification accuracy for these losses, and it can be seen that increases in loss are mirrored by decreases in accuracy.
\autoref{fig:cifar_wnone_losses} shows the loss distribution of the cooperative agents in the presence of the non-cooperative agents from the bottom row of \autoref{fig:cifar_losses}, but with no \cw{} instead of with \wgp{}.
This is different from the top row in the case of the \omniscient{} agent, since for \autoref{fig:cifar_wnone_losses} the non-cooperative agent's knowledge class is still defined relative to \wgp{}, not no weighting.
In \autoref{fig:cifar_losses_n8} we again show the loss distribution of cooperative agents, but we fix the weighting scheme as \wgp{} and the non-cooperative agent class as \cautious{} and use a total number of agents $N=8$ instead of $N=6$.
We then vary the total number of non-cooperative agents $F$, and the maximum number of non-cooperative agents considered possible by the \cw{} scheme, $F_{\text{max}}$

\autoref{fig:cifar_losses} shows that in the absence of any non-cooperative agents the \wgp{} scheme causes a negligible $0.1\%$ increase in shared loss.
In the absence of a weighting scheme, adversaries \faulty{}, \naive{}, \cautious{} and \omniscient{} cause loss increases of $0.024$ ($4.7\%$), $2.934$ ($578\%$), $0.314$ ($61.8\%$) and 0.174 ($34.2\%$) respectively, as seen by comparing the later columns of \autoref{fig:cifar_wnone_losses} with its first.
Comparing the bottom row of \autoref{fig:cifar_losses} with \autoref{fig:cifar_wnone_losses}, we see that the introduction of the \wgp{} scheme reduces these loss increases by $33\%$, $99.7\%$, $97.5\%$ and $15\%$ respectively, almost entirely negating the effects of the \naive{} and \cautious{} adversaries.

\begin{figure}[tb]
  \centering
  \includegraphics{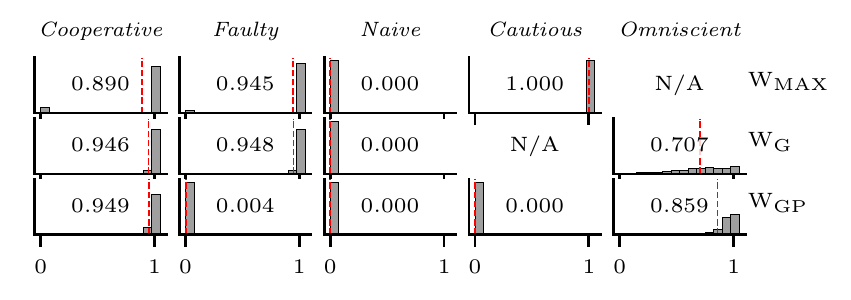}
  \caption{Distributions of confidence weights assigned by the three weighting schemes to all adversaries considered in \autoref{fig:cifar_losses}.}
  \label{fig:cifar_weights}
\end{figure}

\autoref{fig:cifar_weights} is analogous to the lower three rows of \autoref{fig:cifar_losses}, but shows the distribution of confidence weights assigned by the cooperative agents to the non-cooperative agent instead of the loss of the cooperative agents.
Examining these results,
we can see that \wgp{} assigns consistently low weights to adversaries \faulty{}, \naive{} and \cautious{}, but not to \omniscient{}.
We conclude that \wgp{} detects the communication of the \faulty{} adversary but only reduces the extra loss by $33\%$ with respect to no weighting because the unmitigated extra loss is small, whereas \wgp{} only reduces the extra loss of adversary \omniscient{} by $15\%$ because it does not detect its communication as easily.
Also noteworthy in \autoref{fig:cifar_weights} is the comparison between the weights assigned to \faulty{} and \cautious{} adversaries.
While the differences in excess loss with \faulty{} adversaries were small between schemes due to the low impact of the \faulty{} adversary even completely unfiltered, it can clearly be seen that \wgp{} detects and filters out the \faulty{} and \cautious{} adversaries perfectly.
In contrast \wmax{} and \wg{} assign mean confidence weights of $0.945$ and $0.948$ to the \faulty{} adversary respectively, and \wmax{} assigns a mean weight of $1.00$ to the \cautious{} adversary, equivalent to no weighting scheme at all.

We now evaluate alternative weighting schemes \wmax{} and \wg{}, referring to the second and third rows of \autoref{fig:cifar_losses}.
In the presence of adversary \faulty{} \wmax{} increased the extra loss by $54\%$ and \wg{} left it unchanged, as compared to the \wgp{}'s $33\%$ reduction.
Against adversary \naive{}, all schemes were able to reduce the extra loss by $> 99\%$.
The results with \omniscient{} adversaries are more interesting,
and we shall compare the two most effective schemes, \wg{} and \wgp{} in this respect.
Against their respective \omniscient{} adversaries we can see from the final column that \wg{} and \wgp{} incur excess losses with respect to the cooperative baseline loss ($0.508$) of $0.312$ and $0.148$ respectively, a difference of a factor of two in favor of \wgp{}.
We can also contrast these schemes' performance against each other's \omniscient{} adversaries, though this is less important.
Since the \cautious{} agent we used for \wgp{} is actually identical to the \omniscient{} agent of \wg{}, we can see that \wgp{} cuts down the excess loss with this adversary to $0.008$, a reduction of $97.4\%$ with respect to \wg{}.
In contrast, the \wg{} scheme is only able to reduce the excess loss caused by \wgp{}'s \omniscient{} adversary from $0.148$ to $0.106$ (not shown in figures) in comparison to \wgp{} itself, a reduction of only $28\%$, showing that \wgp{} has broader effectiveness across adversaries than \wg{}.

Finally, we see in \autoref{fig:cifar_losses_n8} the loss distribution (corresponding accuracies in \autoref{tab:cifar_accuracy_n8}) when we generalize our method to various numbers of adversarial agents $F$ and a different total number of agents $N$.
The results are for \wgp{} in the presence of the \cautious{} adversary with $N=8$ with varying numbers of adversaries ($F$) and varying maximum $F$ considered by the \cw{} scheme ($F_{\mathrm{max}}$).
We see that so long as $F_{\mathrm{max}} >= F$ the weighting scheme is able to substantially mitigate the effect of the \cautious{} adversaries.
We expect that this consistency of performance across $F$ values generalizes across adversaries, though performance with $F_{\mathrm{max}} > 1$ is unlikely to improve on that of $F_{\mathrm{max}} = 1$ in cases where that original performance was less striking, e.g. with the \omniscient{} adversary.

\begin{table}[tb]
\centering
\caption{Cooperative classification accuracy for \autoref{fig:cifar_losses}.}
\label{tab:cifar_accuracy}
\begin{tabular}{c||c|c|c|c|c}
& \textit{Cooperative} & \textit{Faulty} & \textit{Naive} & \textit{Cautious} & \textit{Omniscient}  \\ \hline\hline
None & $0.750$ & $0.742$ & $0.502$ & $0.549$ & N/A \\\hline
$\mathrm{W}_\mathrm{MAX}$ & $0.720$ & $0.714$ & $0.723$ & $0.563$ & N/A \\\hline
$\mathrm{W}_\mathrm{G}$ & $0.745$ & $0.727$ & $0.733$ & N/A & $0.540$ \\\hline
$\mathrm{W}_\mathrm{GP}$ & $\mathbf{0.762}$ & $\mathbf{0.746}$ & $\mathbf{0.752}$ & $\mathbf{0.735}$ & $\mathbf{0.635}$ \\\end{tabular}
\end{table}

\begin{table}[tb]
\centering
\caption{Cooperative classification accuracy for \autoref{fig:cifar_losses_n8}.}
\label{tab:cifar_accuracy_n8}
\begin{tabular}{c||c|c|c}
& $F_\mathrm{max}=1$ & $F_\mathrm{max}=2$ & $F_\mathrm{max}=3$  \\ \hline\hline
$F=1$ & $0.768$ & $0.746$ & $0.751$ \\\hline
$F=2$ & $0.578$ & $0.748$ & $0.737$ \\\hline
$F=3$ & $0.522$ & $0.536$ & $0.707$ \\\end{tabular}
\end{table}

\subsection{Cooperative Coverage}

In our second case study, we consider the multi-agent coverage path planning problem in a \Gls{rl} setting as described in~\cite{blumenkamp_2020}. In this setting, a non-cooperative agent competes with cooperative agents for coverage. In contrast to the prior case study, the non-cooperative agent's objective \emph{is the same} cooperative agents' objective (i.e., not its negation). Here, the non-cooperative agent is simply \emph{self-interested}, meaning that it does not share the cooperative agents' global reward. In~\cite{blumenkamp_2020}, we showed how this reward structure enabled the self-interested agent to learn manipulative communication policies (benefiting its selfish goal). In the following set of results, we aim to show that our \mfs{} is able to mitigate the impact of its adversarial communication policy.
Specifically, we demonstrate how \cw{} can detect and mitigate manipulative communications for the \naive{} and \omniscient{} case.

\subsubsection{Setup}
\begin{figure}[tb]
    \centering
    \includegraphics[width=0.4\linewidth]{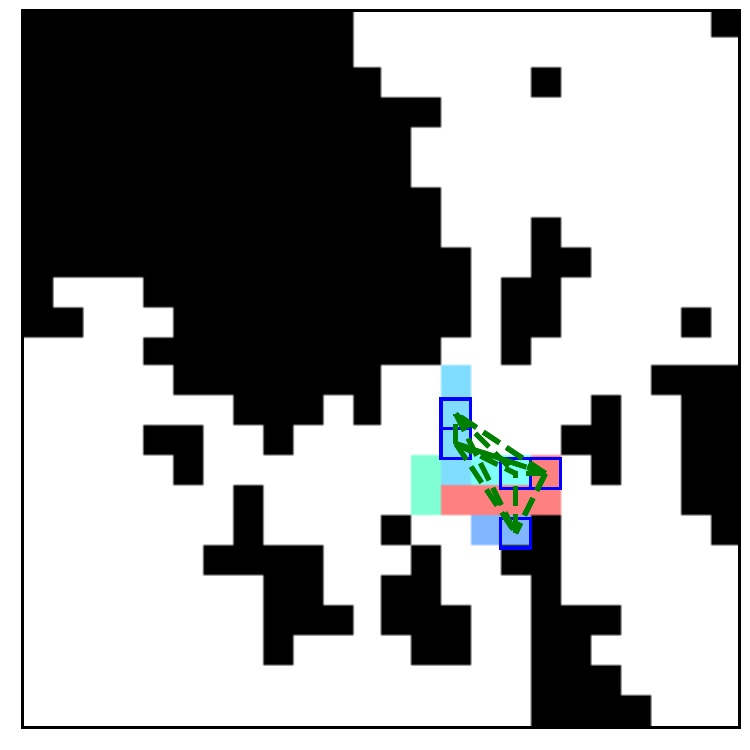}
    \includegraphics[width=0.4\linewidth]{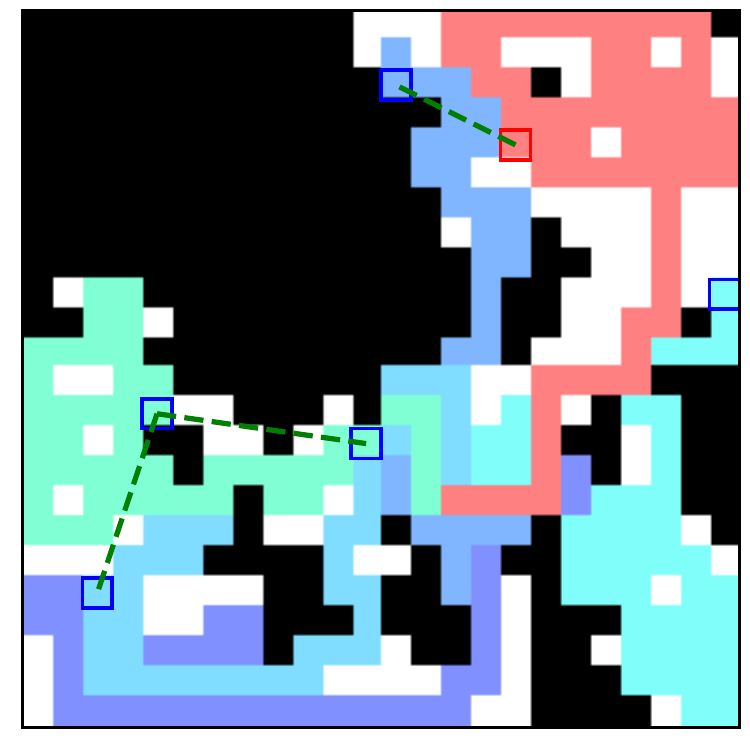}
    \caption{Coverage environment at two different time steps: Six agents (blue and red squares) are placed in a randomly generated grid-world with the goal to cover the free space (white) as quickly as possible. Agents can communicate if they are within communication range $r_c$ (indicated as green dashed lines). The red agent is non-cooperative with an adversarial (self-interested) goal, i.e., to cover as much area as possible, non-cooperatively.} 
    \label{fig:env_coverage}
\end{figure}

We consider a non-convex environment represented as a binary grid world populated with agents that aim to cooperatively and as quickly as possible visit every cell.
The environment of size $24\times24$ is populated with $N=6$ agents, each of which has a local field of view of $16\times16$ pixels separated in two channels representing obstacles and local coverage with a communication radius of $r_c=16$. An overview of the environment can be seen in \autoref{fig:env_coverage}. The agents' communication topology changes over discrete time as the agents move and interact with the environment (i.e., avoid obstacles). 

\subsubsection{Training/Architecture Extensions}

We extend the architecture described in \cite{blumenkamp_2020} as follows both for the self-interested (from now on referred to as non-cooperative) and the cooperative policy: \emph{(i)} replace the encoder with a \Gls{vae} that outputs a multivariate Gaussian \emph{(ii)} add a local encoder of the observation that skips the \Gls{vae} and the \Gls{gnn} to the final action \Gls{mlp} \emph{(iii)} add an \Gls{mlp} transforming the output of the \Gls{vae} before feeding it to the \Gls{gnn} \emph{(iv)} add the \cw{} mechanism that takes into consideration the \Gls{vae} Gaussian of all agents and generates attention weights for the \Gls{gnn} layer.
The training consists of the following steps: First, we train the cooperative policy with the modified architecture, but without training the \Gls{vae} and \Gls{gp}. We use this policy to train the \Gls{vae} and \Gls{gp} independently from the policy by collecting samples of observations and positions of agents. We then train the cooperative policy from scratch with the \Gls{vae} parameters obtained in the previous step while keeping the \Gls{vae} parameters unchanged. This yields a cooperative coverage policy that uses the \Gls{vae} to encode the local observations of each agent, which is shared through the \Gls{gnn} with other agents.

During training, we assume full connectivity for the \cw{} and compute the weights only for the non-cooperative agent. This makes learning manipulative communications for the non-cooperative agents more difficult and therefore emphasizes the success of \cw{}. During evaluation, we use the graph topology resulting from the specified $r_c$ for the \cw{}. Similar as described in \autoref{sec:exp_cifar}, we penalize excessive deviation from $\mathbf{m}_C$ via a squared error loss.

\subsubsection{Experiments}
\label{sec:exp:cov:exp}
To demonstrate the effectiveness of \cw{}, we consider three reference experiments. First, we introduce a baseline policy (with $N=6$) that randomly moves {an agent to a neighboring cell with preference to uncovered cells, without making use of any communication to other agents}.

Secondly, we consider training two different \cooperative{} models with $N=6$ and $F=0$. Both models use the same architecture, but the first model is optimized without pre-trained \Gls{vae}, while the second model uses a pre-trained \Gls{vae}. This means that only the latter can be used with \cw{}. We evaluate the latter with and without \cw{} and with a drop-in median filter, for which we replace the element-wise sum over neighbors in \eqref{eq:gnn_layer} with a median over neighbors. The drop-in median filter corresponds to an alternate (benchmark) \mfs{}.
Additionally, we also perform experiments without any communication.

Lastly, we introduce a single non-cooperative agent so that $N=6$ and $F=1$. We train three models, two that cover the \naive{} case and one that covers the \omniscient{} case. For the \naive{} case, we train similar to the adversarial of \cite{blumenkamp_2020} without \cw{} and evaluate with and without \cw{}. Additionally, we train a model for the \naive{} case with drop-in median. For the \omniscient{} case, we train with \cw{} and evaluate with and without \cw{}.

\subsubsection{Metrics}
For comparison, we consider the percentual global coverage at two fixed time steps $t_0$ and $t_1$, which are, respectively, the number of steps an ideal policy would require to cover the full area if all $N=6$ agents cover a cell at every time-step ($t_0$), or if only a single agent covers a cell at every time step ($t_1$). 
We refer to the coverage (either per-agent or total) at these time steps as $p_0$ and $p_1$, respectively. Our experiments consider $N=6$ agents in a world of size $24\times24$ with $60\%$ coverable space, therefore, $t_0=\lceil \frac{24^2\cdot0.6}{6}\rceil=58$ and $t_1=\lceil \frac{24^2\cdot0.6}{1}\rceil=345$.

\subsubsection{Results}
\begin{figure}[tb]
  \centering
  \includegraphics{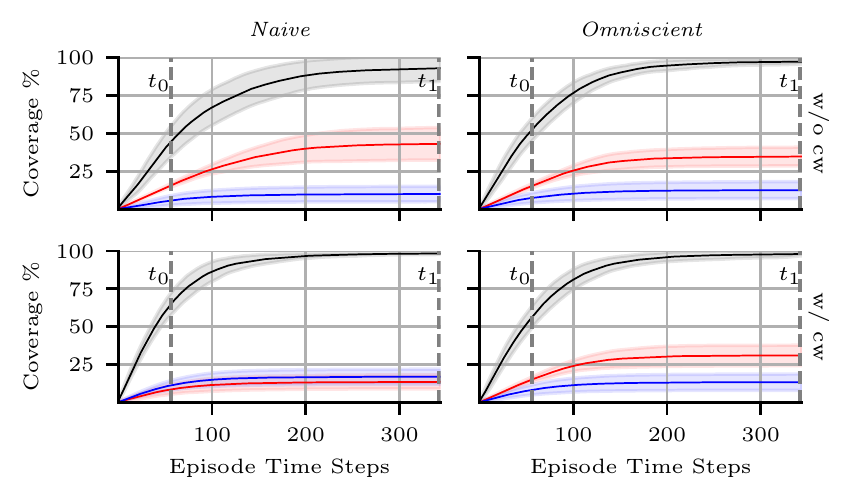}
  \caption{We show the impact of a single non-cooperative agent ($F=1$) on a cooperative team ($N=6$) for the \naive{} case (left column) and \omniscient{} case (right column) as area covered over episode time steps for 100 episodes for the non-cooperative agent (red), mean of cooperative agents (blue) or all agents (black) with a $1\sigma$ standard deviation over each individual agent. The first row shows the results without and the second row with applied \cw{} (cw).}
  \label{fig:cov_overview_adv}
\end{figure}

We summarize the results of all coverage experiments in \autoref{tab:cov_comp_all}. This table separates different training runs, as explained in \autoref{sec:exp:cov:exp}, {with two thin lines or one thick vertical line and different evaluations of the same training run with one thin vertical line. We highlight the best results per experiment group (separated by thick lines) in bold for the $p_0$ and $p_1$ performance both for the cooperative agents and all agents together (higher is better), and non-cooperative agents (lower is better).} \autoref{fig:cov_overview_adv} compares the \naive{} and \omniscient{} case with and without \cw{}.
For the \naive{} case without \cw{}, the non-cooperative agent covers $430\%$ ($p_1$) of the area of an average cooperative agent. The total team coverage performance drops to $64\%$ of that of an entirely cooperative team ($p_0$). In contrast, when applying \cw{} to the cooperative agents in the \naive{} case, the non-cooperative agent manages to cover only $82\%$ of the area an average cooperative agent covers ($p_1$) and the total performance is at $92\%$ of the \cooperative{} performance.

In the \omniscient{} case without \cw{}, the non-cooperative agent covers $269\%$ of the area of an average cooperative agent, or $81\%$ of the non-cooperative coverage for the \naive{} case ($p_1$). 
With applied \cw{}, the performance is similar (i.e., it even drops slightly to $238\%$ and $72\%$, respectively), indicating that the non-cooperative agent was not able to learn better adversarial communications, despite the fact that it was trained with knowledge of the \cw{}. The total team coverage drops to $80\%$ of that of an entirely cooperative team ($p_0$).

\begin{figure*}[tb]
  \centering
  \includegraphics{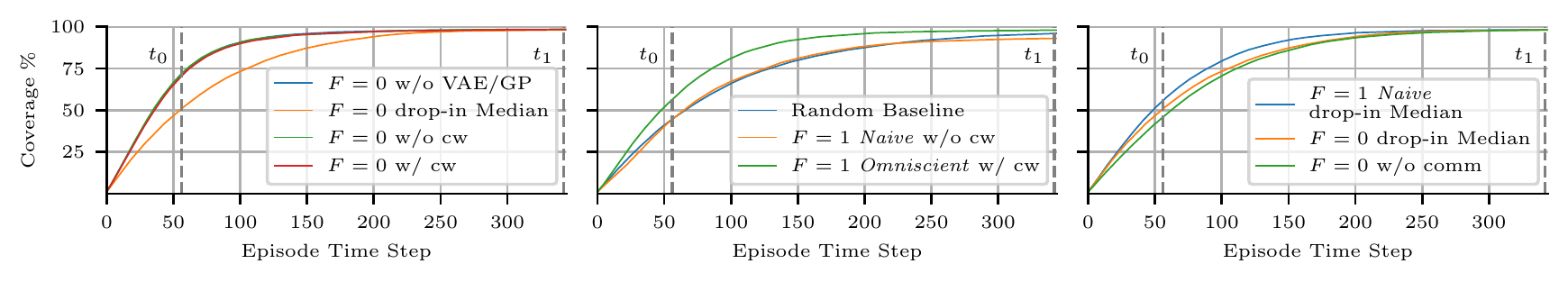}
  \caption{Coverage performance as globally covered area over episode time steps as a comparison of total (all agents) coverage performance for different experiment as mean over 100 episodes. \textit{Left}: We compare a cooperative model trained with a pre-trained \Gls{vae} without and with active \cw{} for all agents to a cooperative model with drop-in median. \textit{Middle}: We compare the impact a single \naive{} and \omniscient{} agent ($F=1$) has on a cooperative team ($N=6$) with and without \cw{} for the cooperative agents. \textit{Right}: Lastly, we compare a cooperative model with drop-in median \omniscient{} with \cw{} performs similar and \naive{} trained with drop-in median with and without communication perform similar.}
  \label{fig:cov_comp_adv}
\end{figure*}

Lastly, we compare the total performance of a subset of trials of \autoref{tab:cov_comp_all} in \autoref{fig:cov_comp_adv}. The plots on the left side compare the impact of \cw{} for the purely \cooperative{} case without non-cooperative agents and with drop-in median. The performance for the \cooperative{} case trained without \Gls{vae} is similar to trained with and evaluated with and without \cw{}. This indicates no negative impact of \cw{} to an entirely cooperative team. In contrast, the drop-in median for the \cooperative{} case performs at $71\%$ of the cooperative trained without \Gls{vae} and therefore significantly affects the cooperative performance. In the middle, we compare the total performance for the \naive{} case without \cw{} to the \omniscient{} case with \cw{}. We see that for the \naive{} case without \cw{}, the performance is similar to the random baseline. In contrast, the \omniscient{} case with \cw{} is at $124\%$ of the \naive{} case without \cw{} or random baseline, indicating a noticeable boost of performance for the cooperative objective to collectively cover the area despite adversarial communications.
The plot on the right side compares the performance for the \naive{} case trained with drop-in median to the \cooperative{} with drop-in median and \cooperative{} without communication between agents. In the \cooperative{} case, the drop-in median performs slightly better than without communication, but significantly worse than any \cooperative{} case with communication. The \naive{} case trained with drop-in median filter performs only slightly better. While the median aggregation scheme filters out adversarial messages, it also filters out most cooperative messages and therefore causes the performance to drop. The \naive{} case with drop-in median filter performs slightly better since the non-cooperative agent can take this into consideration and act greedily, in contrast to an average cooperative agent.

\newcommand{\tb}{\VRule[2pt]}
\begin{table*}[tb]
\centering
\caption{Summary of all coverage results. Best results per group (block between solid separators) in bold.}
\label{tab:cov_comp_all}
\setlength\tabcolsep{3pt} 
\begin{tabular}{cc||c!{\tb} c||c|c|c|c!{\tb}c|c||c!{\tb}c|c}
\multicolumn{2}{c||}{\multirow{4}{*}{\begin{tabular}[c]{@{}c@{}}$N=6$\end{tabular}}} & & \multicolumn{5}{c!{\tb}}{All \cooperative{} ($F=0$)} & \multicolumn{5}{c}{w/ one non-cooperative ($F=1$)} \\
 &  & \multirow{2}{*}{\begin{tabular}[c]{@{}c@{}}Random\\ Baseline\end{tabular}} & \multirow{2}{*}{\begin{tabular}[c]{@{}c@{}}w/o\\ VAE/GP\end{tabular}} & \multicolumn{4}{c!{\tb}}{w/ VAE/GP} & \multicolumn{2}{c||}{\begin{tabular}[c]{@{}c@{}}\naive{} \end{tabular}} & \multirow{3}{*}{\begin{tabular}[c]{@{}c@{}}\naive{} w/ \\ drop-in \\ Median \end{tabular}} & \multicolumn{2}{c}{\omniscient{}} \\
 &  &  &  & \begin{tabular}[c]{@{}c@{}}w/o\\ cw\end{tabular} & \begin{tabular}[c]{@{}c@{}}w/\\ cw\end{tabular} & \begin{tabular}[c]{@{}c@{}}drop-in \\ Median \end{tabular} & \begin{tabular}[c]{@{}c@{}}w/o \\ comm \end{tabular} & \begin{tabular}[c]{@{}c@{}}w/o\\ cw\end{tabular} & \begin{tabular}[c]{@{}c@{}}w/\\ cw\end{tabular} &  & \begin{tabular}[c]{@{}c@{}}w/o\\ cw\end{tabular} & \begin{tabular}[c]{@{}c@{}}w/\\ cw\end{tabular} \\ \hline \hline
  \multirow{2}{*}{\begin{tabular}[c]{@{}c@{}}Cooperative\\ (per agent)\end{tabular}} & $p_0$ & $7.4 \pm 4$ & $11.8 \pm 3$ & $\mathbf{12.0} \pm 3$ & $11.7 \pm 3$ & $8.5 \pm 4$ & $7.6 \pm 5$ & $5.8 \pm 3$ & $\mathbf{11.3} \pm 3$ & $8.4 \pm 4$ & $7.4 \pm 3$ & $\mathbf{8.3} \pm 3$\\ 
& $p_1$ & $16.0 \pm 8$ & $\mathbf{16.4} \pm 4$ & $\mathbf{16.4} \pm 4$ & $\mathbf{16.4} \pm 4$ & $\mathbf{16.4} \pm 8$ & $16.3 \pm 8$ & $10.0 \pm 5$ & $\mathbf{17.0} \pm 5$ & $14.4 \pm 6$ & $12.5 \pm 6$ & $\mathbf{13.4} \pm 5$\\ \hline 
\multirow{2}{*}{\begin{tabular}[c]{@{}c@{}}Non-\\Cooperative\end{tabular}} & $p_0$ & \multirow{2}{*}{N/A} & \multirow{2}{*}{N/A} & \multirow{2}{*}{N/A} & \multirow{2}{*}{N/A} & \multirow{2}{*}{N/A} & \multirow{2}{*}{N/A} & $15.7 \pm 1$ & $\mathbf{8.6} \pm 3$ & $14.0 \pm 2$ & $15.4 \pm 1$ & $\mathbf{15.1} \pm 2$\\ 
 & $p_1$ & & & & & & & $43.1 \pm 11$ & $\mathbf{13.6} \pm 5$ & $26.3 \pm 7$ & $34.7 \pm 6$ & $\mathbf{31.0} \pm 7$\\ \hline 
 Total & $p_0$ & $44.7 \pm 9$ & $70.8 \pm 8$ & $\mathbf{71.9} \pm 7$ & $70.4 \pm 7$ & $50.9 \pm 10$ & $45.9 \pm 11$ & $44.5 \pm 10$ & $\mathbf{64.9} \pm 7$ & $55.9 \pm 8$ & $52.6 \pm 7$ & $\mathbf{56.5} \pm 7$
\end{tabular}
\end{table*}

\section{Discussion and Further Work}
\textbf{Summary.}
In this paper we developed a two-stage probabilistic model of agent observations consisting of a \Gls{gp} prior over the latent vectors of a per-agent auto-encoder, following the general \Gls{aevb} framework~\cite{kingma_2013}.
By using the latent space posteriors produced by the encoder as messages between agents we thereby obtained a probabilistic model describing the joint distribution over agent messages conditional on their relative positions from the \Gls{gp} stage.
This model of messages allows us to formulate and compare hypotheses regarding the generation mechanism for received messages, in particular regarding which agents are being truthful, and thereby assign \textit{confidences} to the messages received from different neighbors.
Having introduced a taxonomy of non-cooperative agents according to their level of knowledge of any countermeasures employed by the cooperative agents, we integrated our \textit{confidences} into a \Gls{gnn} analogously to attention weights in order to filter out potentially harmful messages from non-cooperative neighbors.

\textbf{Discussion.} Our \cw{} method clearly performs well against all adversaries considered that were optimized with imperfect knowledge of it.
In both of our experiments we find that it can easily detect the communication of adversaries, and, having done so, cause the cooperative agents to disregard it and prevent substantial harm to their individual performance.
Indeed, in our second experiment the remaining decrease in collective performance relative to purely cooperative agents when using \cw{} against a single \naive{} adversary is due to poor performance of the adversary individually at achieving the collective goal via its actions, rather than any effect it has via communication.

We have seen that our method also has desirable performance characteristics outside of this ideal set of scenarios, for example by having a negligible impact on the performance of the cooperative agents if there is in fact no adversary present.
The performance in the presence of an \omniscient{} adversary is more subtle but still impressive.
It is to be noted that no \mfs{} with a negligible impact on purely cooperative agents can reliably detect and filter out the communication of all possible adversaries since a sufficiently capable adversary can produce communication which is arbitrarily similar to actual cooperative communication if necessary.
The strength of a \mfs{} in the presence of a competent \omniscient{} adversary therefore should be measured in terms of the damage caused to the cooperative agents objective by the strongest attack which can bypass it.
This indirect reduction in damage is clearly seen in both experiments, with overall collective performance in the second experiment being reduced by only $21\%$ in an attack optimized against our \mfs{} compared to $37\%$ against no filtering.
This effect was even stronger in the first experiment, where we observed worst case loss increases for the cooperative agents of $29\%$ with \cw{} as opposed to at least $678\%$ without.

We have also seen that our method compares favorably to other alternatives which we have considered.
In the case of the alternative methods considered in the first experiment, our method is superior at suppressing the impact of a wider range of adversaries, while, in the case of the median filtering based method considered in the second experiment, our method has a far smaller negative impact on the performance of the cooperative agents when there are no adversaries present.

\textbf{Further Work.}
Our choice of \Gls{gp} kernel function was chosen to maximize expressiveness while being clearly stationary, but introduced some inconveniences in training due to occasional invalid covariance matrices.
These inconveniences could be avoided if stationarity was not required (e.g. because the agents had access to their absolute position), or by enforcing this stationarity approximately, e.g. by randomly translating the training data.

Our method would permit extensions to take advantage of information which we have assumed (in \autoref{sec:problem}) to not be available.
Examples would include considering all previous communication observed from neighboring agents for the purpose of determining confidence weights, or checking for consistency between the behavior of other agents and their claimed observations given that the cooperative policy is known.
Either of these extensions would substantially reduce the strategies available to an \omniscient{} adversary and further reduce the potential damage caused by such an adversary.

\appendices
\section{Weighting Details}
\label{app:weight_derivation}
We have $Q(\mathbf{Z} | \mathbf{O}; \phi)$ and $P(\mathbf{Z} | \mathbf{X}; \psi, h)$ and wish to compare hypotheses $h$. $\mathbf{O}$ is the same for all hypotheses, though it is unobserved.
We compare $\log P(\mathbf{O} | \mathbf{X}; \psi, \theta, h)$ via \eqref{eq:varbayes}.
It can be seen that the third term is constant across hypotheses, and we can calculate the second.
The first term is intractable, but to the extent that $D_{KL}(Q(\mathbf{O} | \mathbf{Z}; \phi) || P(\mathbf{Z} | \mathbf{O}; \psi, \theta, h))$ is approximately constant across hypotheses we can still compare the various $h$.
Making this assumption we obtain the result
\begin{multline}
    \log P(\mathbf{O} | \mathbf{X}; \psi, \theta, h) \approx \\
    -D_{KL} (
    Q(\mathbf{Z} | \mathbf{O}; \phi) ||
    P(\mathbf{Z} | \mathbf{X}; \psi, h)
    ) + C
\end{multline}
where $C$ is constant across hypotheses $h$.

The assumption is most clearly justified for the $H_0$ and $H_1$ single-agent hypotheses, since $Q(\mathbf{Z} | \mathbf{O}; \phi)$ has been optimized to minimize this term during training, with the only difference in $P(\mathbf{Z} | \mathbf{O}; \psi, \theta, h)$ being the number of points drawn from the \Gls{gp} on which to condition, which $Q(\mathbf{Z} | \mathbf{O}; \phi)$ ignores anyway.
It is less clearly justified for $H_2$, but this hypothesis is already fairly crude, being a uniform distribution in latent space, and we find that it still gets assigned a high marginal likelihood relative to other hypotheses in the case of extreme $\mathbf{z}$ values.

\section{Kernel Function Implementation}
\label{app:kernel_implementation}

Let us consider a neighborhood of $N$ agents, whose latent spaces each have $Z$ dimensions. We wish to define a kernel function which produces $N \times Z$ dimensional covariance matrices $\mathbf{C}$ describing the correlations of all of these latent variables as a function of the relative positions of these N agents. Let the indices $i$ and $j$ index over agents and the indices $k$ and $l$ index over latent variables, that is, let $c_{ijkl}$ refer to the covariance of the $k$th and $l$th latent variables of the $i$th and $j$th agents respectively.

Since we require our \Gls{gp} to be stationary, the $Z$ dimensional covariance submatrix $\mathbf{c}_{ij}$ will always be the same if $i=j$. Since this is also the covariance matrix of the latent variables of an individual agent considered alone, we let these matrices equal $\gamma \mathbf{I}$ similarly to a simple \Gls{vae}. This means that our $N \times Z$ covariance matrix now has the form, shown here for $N=3$:

$$\begin{bmatrix}
\gamma \mathbf{I} & \mathbf{c}_{01}^T & \mathbf{c}_{02}^T \\
\mathbf{c}_{01} & \gamma \mathbf{I} & \mathbf{c}_{12}^T \\
\mathbf{c}_{02} & \mathbf{c}_{12} & \gamma \mathbf{I} \\
\end{bmatrix}$$

We wish to define the covariances of $\mathbf{c}_{ij}$ by a differentiable function of relative position $\mathbf{x}_{ij} = \mathbf{x}_j - \mathbf{x}_i$ involving a neural network. We also require that these covariances form a valid covariance matrix for every pair of agents when combined with the intra-agent covariances $\gamma \mathbf{I}$.
We start by defining a NN function $L(\mathbf{x}_{ij})$ from relative positions to $2Z \times M$ matrices, where $M \leq 2Z$, parameterized as an \Gls{mlp}. $LL^T$ is then guaranteed to be a valid $2Z$ dimensional covariance matrix, whose submatrices we shall denote as
$$\begin{bmatrix}
\mathbf{t}_{ij} & \mathbf{m}_{ij}^T \\
\mathbf{m}_{ij} & \mathbf{b}_{ij} \\
\end{bmatrix}$$
For every off-diagonal element $t_{ijkl}, k \neq l$ of $\mathbf{t}_{ij}$ which is non-zero we can set it to zero by adding a matrix $\mathbf{n}$ whose elements are defined as $n_{pq} = -m_{ijkl}$ if $p=i$ and $q=j$, $n_{pq} = |m_{ijkl}|$ if $p=q=i$ or $p=q=j$ and $0$ otherwise.
This process preserves the validity of the overall $2Z$ dimensional matrix as a covariance matrix since these matrices $\mathbf{n}$ are positive semi-definite.
Likewise, we can increase the values of each diagonal element of the modified $LL^T$ to equal the maximum element of the diagonal while retaining validity.
We can thereby obtain a modified valid $2Z$ dimensional covariance matrix with the form
$$\begin{bmatrix}
\beta_{ij} \mathbf{I} & \mathbf{m}_{ij}^T \\
\mathbf{m}_{ij} & \beta_{ij} \mathbf{I}\\
\end{bmatrix}$$
for some value $\beta_{ij}$.
We can explicitly calculate $\beta_{ij}$ as
$\beta_{ij} = \max (\beta_{ij_{t}}, \beta_{ij_{b}})$
where
$\beta_{ij_{t}} = \max_k \sum_l |t_{ijkl}|$
and
$\beta_{ij_{b}} = \max_k \sum_l |b_{ijkl}|$
.
This covariance matrix can now be made compatible with the intra-agent covariance matrices of $\gamma \mathbf{I}$ by multiplying the result by $\frac{\gamma}{\beta_{ij}}$, giving us:
\begin{equation}
    \mathbf{c}_{ij} = \frac{\gamma}{\beta_{ij}} \mathbf{m}_{ij}
\end{equation}

This process guarantees that a $\mathbf{C}$ generated for any pair of agents is a valid covariance matrix, and is dependent only on relative position, but does not guarantee validity for $N \geq 3$. For this we rely on the trained function becoming valid for almost all input once it converges while being trained on pairs only, an effect which we do in fact observe. (To make our method invariant in permutation of the agents, we take the further step of symmetrizing and use $\frac{1}{2} (\mathbf{c}_{ij} + \mathbf{c}_{ji}^T)$ in place of just $\mathbf{c}_{ij}$, where $\mathbf{c}_{ji}^T$ is calculated similarly to $\mathbf{c}_{ij}$ but with the relative position negated.)

\section*{Acknowledgment}
We gratefully acknowledge the support of ARL grant DCIST CRA W911NF-17-2-0181. A. Prorok was supported by the Engineering and Physical Sciences Research Council (grant EP/S015493/1). J. Blumenkamp was supported in part through an Amazon Research Award.

\addtolength{\textheight}{-16cm}

\bibliographystyle{IEEEtran}
\bibliography{IEEEabrv, bibliography}

\end{document}